\documentclass{article}

\usepackage[english]{babel}
\usepackage{float}
\usepackage[numbers]{natbib}

\usepackage[a4paper,top=2cm,bottom=2cm,left=3cm,right=3cm,marginparwidth=1.75cm]{geometry}

\usepackage[utf8]{inputenc} 
\usepackage[T1]{fontenc}    
\usepackage{hyperref}       
\usepackage{url}            
\usepackage{booktabs}       
\usepackage{amsfonts}       
\usepackage{nicefrac}       
\usepackage{microtype}      
\usepackage{xcolor}         

\usepackage{enumitem}
\usepackage[linesnumbered,ruled,vlined]{algorithm2e}

\usepackage{amsmath}
\usepackage{wrapfig}
\usepackage{tikz}
\usetikzlibrary{shapes.geometric, arrows.meta, positioning}
\usetikzlibrary{backgrounds, fit, calc}
\usepackage{subcaption}
\usepackage[symbol]{footmisc}

\usepackage{amssymb}
\usepackage{siunitx}
\PassOptionsToPackage{hyphens}{url}\usepackage{hyperref}
\usepackage{cleveref}
\usepackage[utf8]{inputenc}
\usepackage[right]{lineno}
\usepackage{csquotes}
\usepackage{booktabs}
\usepackage{longtable}
\usepackage{adjustbox}
\usepackage{array}
\usepackage{url}
\usepackage{titlesec}
\usepackage{authblk}
\usepackage{xcolor} 

\titleformat{\subsection}
  {\mdseries\itshape\large} 
  {\thesubsection}{1em}{} 

\hypersetup{
  colorlinks=false,
  pdfborder={0 0 0},
}
\usepackage[english]{babel}
\crefformat{figure}{#2Figure~#1#3}
\Crefformat{figure}{#2Figure~#1#3}
\crefformat{table}{#2Table~#1#3}
\Crefformat{table}{#2Table~#1#3}
\crefformat{section}{#2Section~#1#3}
\Crefformat{section}{#2Section~#1#3}

\author{Vibhhu Sharma\thanks{\texttt{vs552@cornell.edu}}}
\author{Thorsten Joachims}
\author{Sarah Dean}
\affil{Cornell University}

\title{Do LLMs Favor LLMs? Quantifying Interaction Effects in Peer Review}

\usepackage{amsthm}
\usepackage{graphicx}
\usepackage{subcaption}
\usepackage{booktabs}

\usepackage{amsmath}
\usepackage{macros}

\begin{document}

\maketitle

\begin{abstract}
\noindent There are increasing indications that LLMs are not only used for producing scientific papers, but also as part of the peer review process. In this work, we provide the first comprehensive analysis of LLM use across the peer review pipeline, with particular attention to interaction effects: not just whether LLM-assisted papers or LLM-assisted reviews are different in isolation, but whether LLM-assisted reviews evaluate LLM-assisted papers differently. In particular, we analyze over 125,000 paper-review pairs from ICLR, NeurIPS, and ICML. We initially observe what appears to be a systematic interaction effect: LLM-assisted reviews seem especially kind to LLM-assisted papers compared to papers with minimal LLM use. However, controlling for paper quality reveals a different story: LLM-assisted reviews are simply more lenient toward lower quality papers in general, and the over-representation of LLM-assisted papers among weaker submissions creates a spurious interaction effect rather than genuine preferential treatment of LLM-generated content. By augmenting our observational findings with reviews that are fully LLM-generated, we find that fully LLM-generated reviews exhibit severe rating compression that fails to discriminate paper quality, while human reviewers using LLMs substantially reduce this leniency. Finally, examining metareviews, we find that LLM-assisted metareviews are more likely to render accept decisions than human metareviews given equivalent reviewer scores, though fully LLM-generated metareviews tend to be harsher. This suggests that meta-reviewers do not merely outsource the decision-making to the LLM. These findings provide important input for developing policies that govern the use of LLMs during peer review, and they more generally indicate how LLMs interact with existing decision-making processes.
\end{abstract}

\section{Introduction}
\label{sec:intro}
Since ChatGPT's release in late 2022, LLMs have become a fixture in research workflows, with 81$\%$ of researchers now incorporating them into some aspect of their work  (\citet{liao2024llmsresearchtoolslarge}). Recent work has shown evidence of an increasing trend in LLM use by both authors and reviewers at conferences: \citet{liaodetection} found that up to $17\%$ of reviews at top machine learning conferences show evidence of LLM modification, while \citet{liang2024mappingincreasingusellms} observed that the abstracts of up to $17.5\%$ of recent Computer Science papers showed evidence of LLM use---a significantly higher number than that for papers in other domains. Anecdotally, some researchers have attributed subpar reviews they have received in recent years to LLM use, indicating a general sentiment of distrust in the use of LLMs in peer review. 

\begin{figure}[t]
\centering
\begin{subfigure}{0.32\textwidth}
    \centering
    \includegraphics[width=\textwidth]{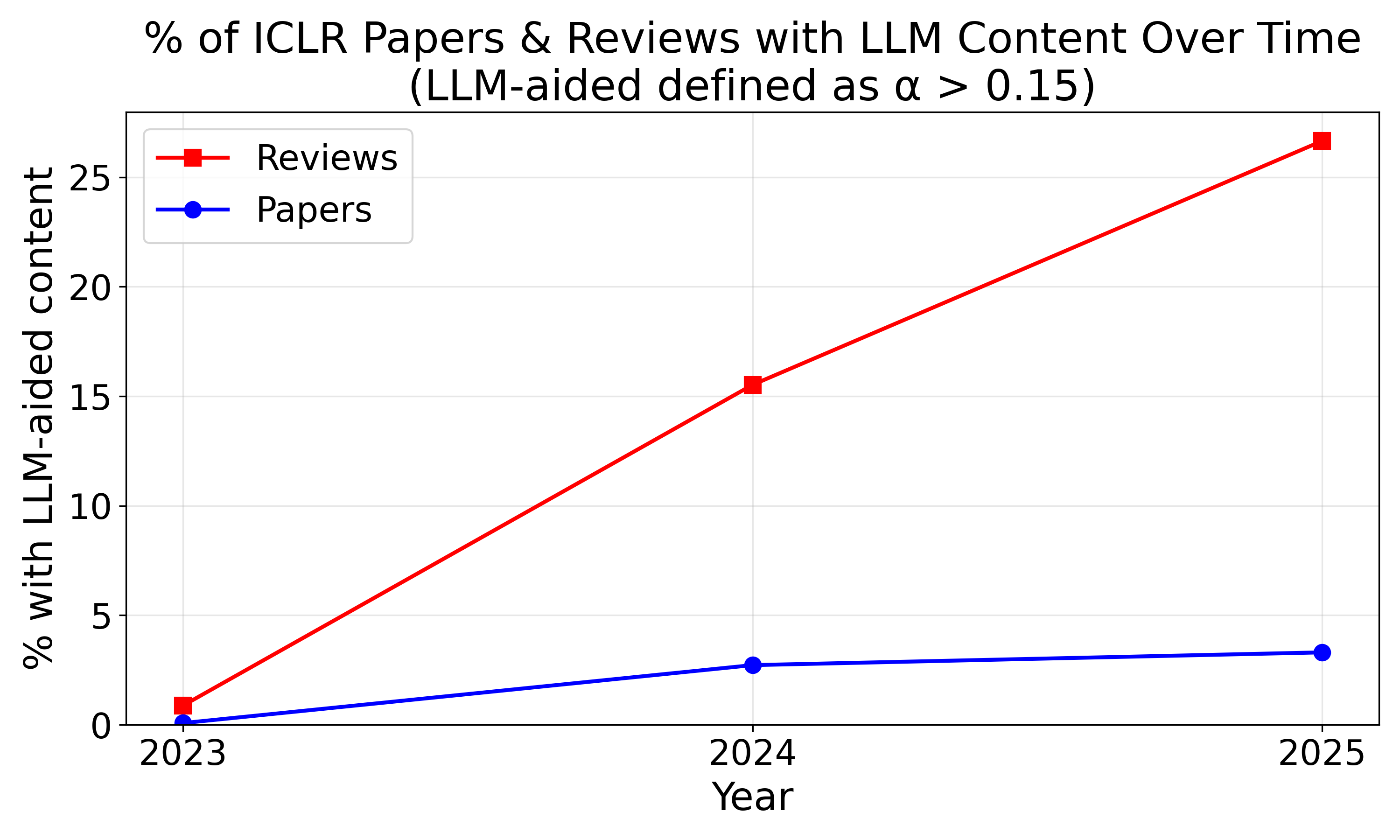}
    \caption{LLM-Aided Papers \& Reviews}
    \label{fig:increaseinllmuse}
\end{subfigure}
\hfill
\begin{subfigure}{0.32\textwidth}
    \centering
    \includegraphics[width=\textwidth]{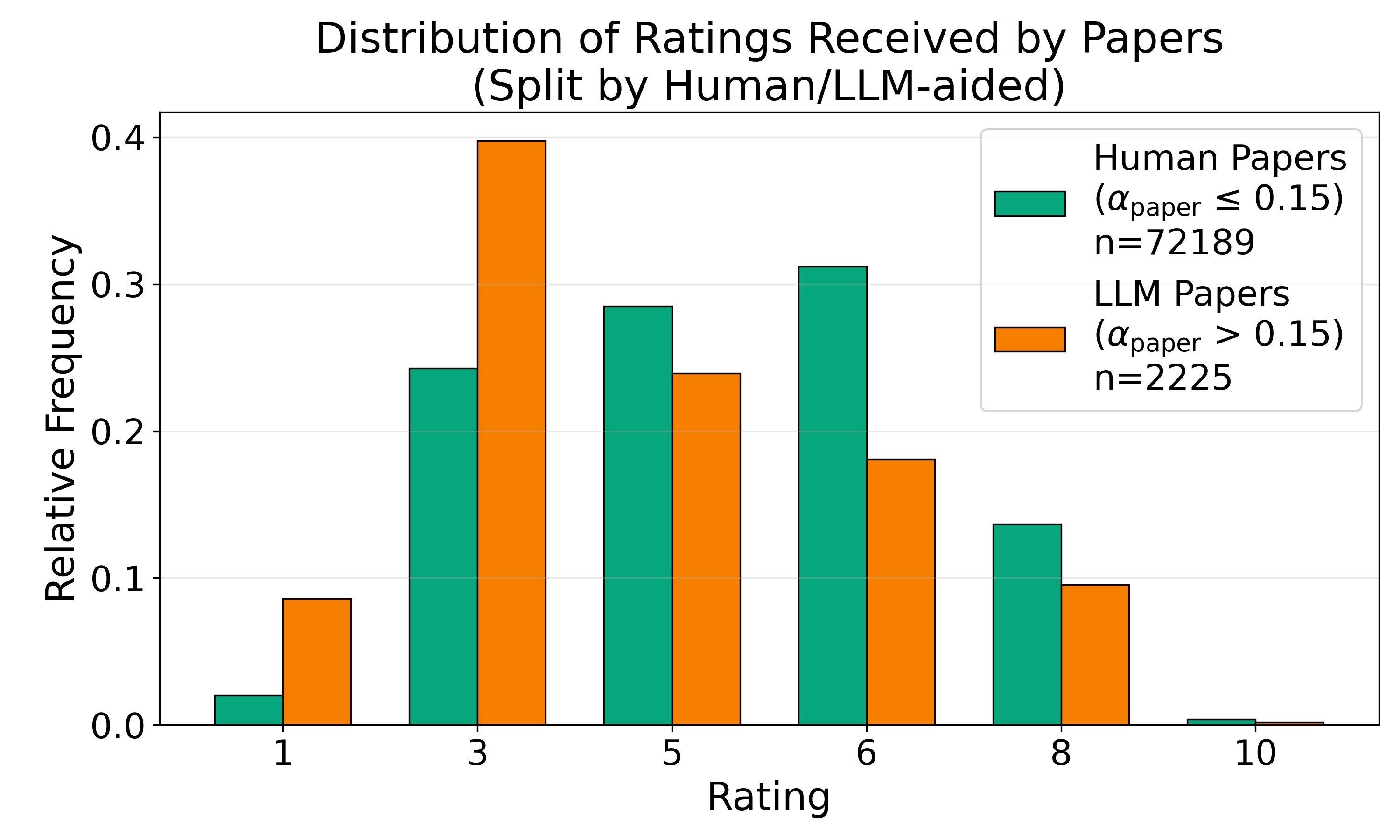}
    \caption{Distribution of paper ratings}
    \label{fig:paperratingdist}
\end{subfigure}
\hfill
\begin{subfigure}{0.32\textwidth}
    \centering
    \includegraphics[width=\textwidth]{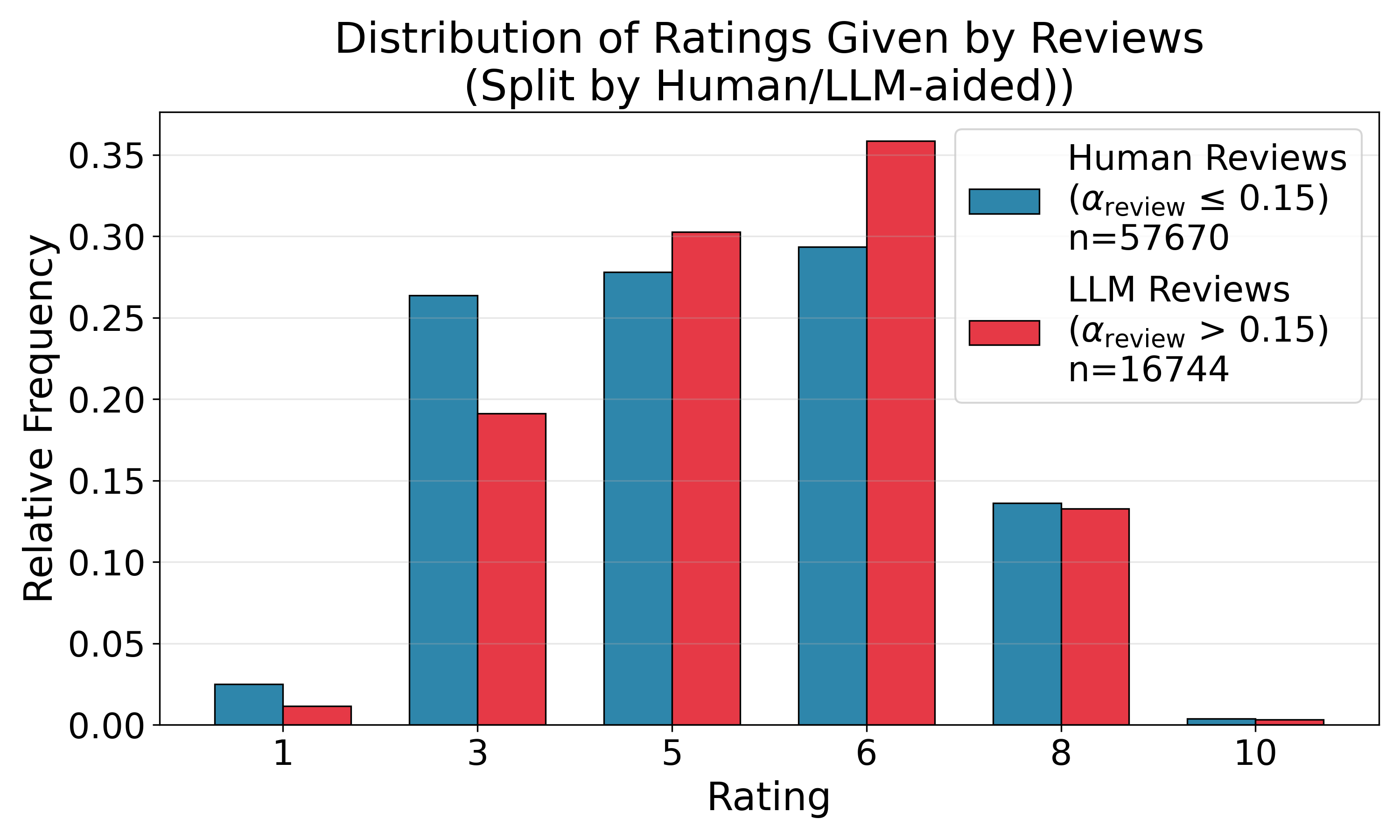}
    \caption{Distribution of review ratings}
    \label{fig:reviewratingdist}
\end{subfigure}
\caption{(a) LLM use trends over time at ICLR. (b-c) Rating distributions for human vs. LLM-aided papers and reviews for ICLR 2024 and 2025 submissions. LLM-aided papers receive a higher distribution of lower scores while LLM-aided reviews more frequently give middling scores relative to their fully human counterparts. Here LLM-aided means the text showed sufficient similarity to LLM vocabulary ($\alpha>0.15$ in Section \ref{sec:llmdetectionmethod}). }
\label{fig:llmandhumanratingdistributions}
\end{figure}

Figure \ref{fig:increaseinllmuse} shows the year by year increase in LLM use, in both papers and reviews submitted to the International Conference on Learning Representations (ICLR), by using the technique from \citet{liaodetection} to estimate the fraction $\alpha$ of LLM generated text in a document (described in Section \ref{sec:data}). Notably, LLM use is substantially more common in writing reviews than in writing papers: in the latest edition of the conference, $3.31\%$ of papers and $26.65\%$ of reviews show substantial LLM modification (detected fraction over $\alpha_{\text{threshold}}= 15\%$). The increase in LLM use in review writing is particularly steep, almost doubling in magnitude in a year. 

Both LLM-assisted reviews and LLM-assisted papers show interesting differences from those that do not have detected LLM assistance (henceforth referred to as "human"). LLM-assisted papers are frequently given lower scores (Fig.~\ref{fig:paperratingdist}), with `3' being the most commonly assigned quality score, relative to a mode of `6' for human-written papers. Meanwhile LLM assisted reviews concentrate toward assigning intermediate ratings and give fewer harsh ratings (Fig.~\ref{fig:reviewratingdist}), suggesting a tendency to hedge rather than provide extreme assessments on either side.

An interesting question that naturally arises from LLM use on both sides of the peer review process regards the existence of a systemic interaction effect between LLM/Human papers and LLM/Human Reviews. More directly, \textit{does a human reviewer behave differently when presented a fully human paper vs an LLM-assisted paper?} Similarly, \textit{does an LLM-assisted reviewer behave differently when presented an LLM-assisted paper vs. a fully human paper?} Interestingly, ICML has instituted a new policy for their 2026 edition (\citet{icml2026llmpolicy}) requiring both reviewers and authors to declare either conservative or permissive LLM use, with reviewer-author pairing done based on these declarations. If any sort of systematic bias does exist in LLM-LLM interactions within peer review, such a policy could inadvertently amplify it. Moreover, authors could potentially game the system if they understand these interaction effects and their implications for review outcomes.  

As an initial datapoint, Table \ref{tab:iclrfullsummary} divides (paper, review) pairs into 4 quadrants based on LLM usage---both human-written, human paper with LLM-assisted review, LLM-assisted paper with human review, and both LLM-assisted---and presents the mean and standard deviations of the scores for these pairs. We find that LLM-assisted papers receive lower ratings on average while LLM-assisted reviews give higher ratings. Both human and LLM-assisted papers receive higher scores when reviewed by an LLM-assisted reviewer, but notably the magnitude differs: \textit{a 0.25-point jump for human papers versus 0.63 points for LLM-assisted papers.} Meanwhile, reviewers show higher confidence for LLM-assisted papers, with human reviewers slightly more confident than LLM-assisted reviewers.

\begin{table}[t]
\centering
\small
\begin{adjustbox}{max width=\textwidth}  
\begin{tabular}{|l|c|c|c|c|c|}
\hline
\textbf{Quadrant }\textit{(Paper, Review)} & \textbf{Rating} & \textbf{Confidence} & \textbf{Presentation} & \textbf{Contribution} & \textbf{Soundness} \\
\hline
(Human, Human) & $5.12 \pm 1.74$ & $3.66 \pm 0.80$ & $2.61 \pm 0.76$ & $2.35 \pm 0.70$ & $2.59 \pm 0.70$ \\
\hline
(Human, LLM) & $5.37 \pm 1.55$ & $3.61 \pm 0.82$ & $2.72 \pm 0.67$ & $2.46 \pm 0.67$ & $2.68 \pm 0.63$ \\
\hline
(LLM, Human) & $4.15 \pm 1.88$ & $3.85 \pm 0.80$ & $2.30 \pm 0.84$ & $2.04 \pm 0.74$ & $2.23 \pm 0.76$ \\
\hline
(LLM, LLM) & $4.78 \pm 1.87$ & $3.80 \pm 0.79$ & $2.48 \pm 0.76$ & $2.29 \pm 0.75$ & $2.48 \pm 0.72$ \\
\hline
\end{tabular}
\end{adjustbox}
\caption{Summary Statistics for (Paper, Review) pairs from ICLR 2024 and 2025 combined. (Paper, Review) pairs are divided into 4 quadrants based on the LLM status of the paper and the review. $\alpha$ value of 0.15 is used as a threshold: papers and reviews are classified as LLM-assisted if they have $\alpha>0.15$. Cell values are presented as Mean $\pm$ Std Dev.}
\label{tab:iclrfullsummary}
\end{table}

These observations suggest potential systematic effects worth exploring, though summary statistics alone cannot verify this due to confounding. It is wholly possible that the papers in one quadrant are fundamentally different papers from the papers in other quadrants, making rating comparisons for these papers moot. This necessitates a causally grounded analysis to eliminate paper-level confounding factors. 

Motivated by the above observations, our paper uses observational data (paper text, metadata, reviews, scores and metareviews) from top machine learning conferences over the last 3 years, in addition to synthetic data generated by prompting LLMs, to answer the following key research questions:
\begin{enumerate}
    \item \textit{Are LLM-aided reviews kinder to LLM-aided papers} on average?
    \item \textit{Does the kindness of LLM-aided reviews to LLM-aided papers vary with paper quality?}
    \item \textit{How does an LLM-aided review differ from a fully LLM-generated review}?
    \item \textit{How does LLM-assistance in metareviewing relate to a paper's final acceptance decision?}
\end{enumerate}
Our work differs from prior studies in several key ways. Unlike research that examines LLM use in papers or reviews in isolation, we provide the first comprehensive analysis of interaction effects between LLM use on both sides of the peer-review process. We combine observational analysis of over 125,000 paper-review pairs from multiple top-tier conferences with controlled synthetic experiments to disentangle genuine LLM effects from the effect of the human in the loop. Our findings on metareview leniency, the quality-dependent nature of LLM review bias, and the moderating role of human-in-the-loop mediation are novel contributions to understanding how AI tools are reshaping research dissemination and decision making.

\section{Related Work}

The rapid adoption of LLMs in research has motivated the development of detection methods to quantify their use. \citet{liaodetection} introduced a statistical approach finding that up to 17\% of reviews at top ML conferences showed significant LLM modification, while \citet{liang2024mappingincreasingusellms} documented 17.5\% of recent Computer Science paper abstracts showed evidence of LLM use. However, they do not study how these distinct forms of LLM use interact with each other.

Several studies have examined how LLM-generated reviews differ from human reviews. \citet{zhu2025reviewerllmbiasesdivergence} identified systematic biases in LLM reviewers, including inflated ratings for weaker papers and reduced sensitivity to paper quality. The susceptibility of LLM reviews to manipulation has been documented by \citet{ye2024yetrevealingrisksutilizing} and \citet{keuper2025promptinjectionattacksllm}, who demonstrated successful prompt injection attacks. However, these studies do not explicitly account for LLM-aided reviews. \citet{KOCAK2025100018} advocate for LLMs as complementary tools rather than replacements for human judgment. 

Beyond peer review, several studies have investigated whether LLMs exhibit bias toward their own outputs. \citet{zheng2023judgingllmasajudgemtbenchchatbot} adopt the term `self-enhancement bias' from social cognition literature. \citet{favor1} found evidence that LLMs preferentially rate their own generated text higher, while \citet{favor2} extended this to binary choice scenarios, showing LLM-based assistants consistently prefer LLM-presented options. This is an important motivation for our work, and we argue that the setting of peer review serves as a valuable testbed for understanding the implications of two-sided LLM use in practice. OpenReview's transparency policies provide rich, structured data that allows us to study authentic high-stakes author-reviewer interactions where both parties may have used LLMs.

\section{Datasets}
\label{sec:data}
We conduct experiments on paper and review data from OpenReview for ICLR (2024, 2025), the Conference on Neural Information Processing Systems (NeurIPS)  (2024, 2025) and the International Conference on Machine Learning (ICML) (2025). These conferences conduct double-blind peer review through OpenReview. Each paper typically receives 3-4 reviews from reviewers assigned based on expertise and preferences. After initial reviews, authors submit rebuttals, followed by a discussion phase between authors and reviewers, where reviewers can observe co-reviewers' assessments. Area chairs then write metareviews recommending decisions to the program committee.
ICLR is unique because of its policy to keep public OpenReview pages for every paper after the conference, irrespective of its acceptance or rejection at the conference. The papers from NeurIPS and ICML are largely those that were accepted, as only a small fraction of researchers chose to keep the OpenReview pages for their rejected papers publicly accessible. 
For each available paper, we collect full paper and review text, the ratings assigned to the paper, and paper metadata (paper area and decision). This is an average of 6.5k papers and 25k reviews per conference (full data statistics are included in Appendix \ref{sec:appdata}).

\emph{LLM Detection.} To detect LLM-generated text in papers and reviews, we use the statistical method from \citet{liaodetection, liang2024mappingincreasingusellms}, which models LLM-generated and human text as distinct distributions. The method assigns each vocabulary token a probability under each distribution based on its frequency in representative training corpora. It then estimates the fraction $\alpha$ of LLM-generated text via maximum likelihood estimation: the method posits any text to have been drawn from a weighted distribution of LLM text and Human text, and returns $\alpha$, which is the mixture weight of the LLM text distribution in the solution to the MLE problem (see Appendix \ref{sec:llmdetectionmethod} for details). In this way, $\alpha$ depends on the bag-of-words representation of each paper or review. We set $\alpha_{\text{threshold}}=0.15$, meaning that any text with $\alpha>\alpha_{\text{threshold}}$ is classified as LLM-aided and human otherwise. This method offers key advantages: ease of use, fast inference and robustness to text that is merely "LLM aided" as compared to fully LLM generated. The reference LLM distribution we use is based on GPT-3.5 and GPT-4o outputs, which were the predominant models during our study period. This approach may underestimate usage of other contemporaneous models (e.g., Claude, Gemini), though GPT-4o's dominant market share at the time suggests this is a minor limitation.
\section{Results}
We now turn to our core empirical questions, analyzing interaction effects between LLM use in papers and reviews. Our analysis proceeds in four stages. First, we examine whether LLM reviews exhibit differential kindness to LLM papers in the full ICLR dataset.
Then, we investigate how this varies by paper quality across venues.
Next, we disentangle the behavior of fully LLM-generated reviews from LLM-assisted reviews found in the wild.
Finally, we examine whether LLM assistance in metareviews affects their propensity to render accept decisions.
\subsection{Are LLM-aided reviews kinder to LLM-aided papers?}
\label{sec:plainiclr}

While the summary statistics in Table \ref{tab:iclrfullsummary} suggest the existence of a differential kindness exhibited by LLM-aided reviews to LLM-aided papers, these statistics cannot paint the whole picture by themselves. Comparing average ratings across quadrants implicitly assumes papers in each quadrant are comparable. However, several sources of bias may undermine this assumption: certain research areas may have both higher LLM adoption rates and different baseline rating distributions, while less experienced authors may be more likely to use LLMs while also producing lower quality work for reasons unrelated to LLM use. This is a classic case of confounding, where aggregate comparisons can be misleading.

\subsubsection{Regression with area correction}
\label{sec:reg}
\begin{figure}[t]
\centering
\begin{tikzpicture}[scale=0.7,
    treatment/.style={circle, draw, minimum size=1cm, fill=green!20},
    covariate/.style={circle, draw, minimum size=1cm, fill=blue!20},
    outcome/.style={circle, draw, minimum size=1cm, fill=orange!20},
    confounder/.style={ellipse, draw, minimum width=1.5cm, minimum height=0.8cm, fill=gray!20},
    unobs/.style={ellipse, draw, dashed, minimum width=1.2cm, minimum height=0.8cm, fill=red!10},
    arrow/.style={-{Stealth}, thick},
    dashed arrow/.style={-{Stealth}, thick, dashed}
]
\node[confounder] (Xarea) at (3, 1.8) {$X_i^{\text{area}}$};
\node[covariate] (Xllm) at (3, 0) {$X_i^{\text{LLM}}$};
\node[treatment] (T) at (6, 0) {$T_{ij}$};
\node[outcome] (Y) at (9, 0) {$Y_{ij}$};
\node[unobs] (U) at (9, 1.8) {$U$};

\node[above=0.1cm of Xarea, font=\footnotesize] {Paper Area};
\node[left=0.1cm of Xllm, font=\footnotesize] {LLM-Assisted Paper};
\node[below=0.25cm of T, font=\footnotesize, align=center] {Treatment\\(LLM-Assisted Review)};
\node[right=0.1cm of Y, font=\footnotesize, align=center] {Outcome\\(Rating)};
\node[above=0.1cm of U, font=\footnotesize] {Unobserved};

\draw[arrow] (Xarea) -- (Xllm);
\draw[arrow] (Xarea) to [bend left=25] (T);
\draw[arrow] (Xarea) to [bend left=20] (Y);
\draw[arrow] (Xllm) -- (T);
\draw[arrow] (Xllm) to [bend right=25] (Y);
\draw[arrow] (T) -- (Y);
\draw[dashed arrow] (U) -- (T);
\draw[dashed arrow] (U) -- (Y);
\end{tikzpicture}
\caption{Causal diagram for LLM use in peer review. $T_{ij}$: treatment (LLM-assisted review). $Y_{ij}$: outcome (rating). $X_i^{\text{LLM}}$: paper LLM use. $X_i^{\text{area}}$: paper area (confounder). Solid arrows are hypothesized causal effects; dashed arrows are potential unobserved confounding.}
\label{fig:causal}
\end{figure}
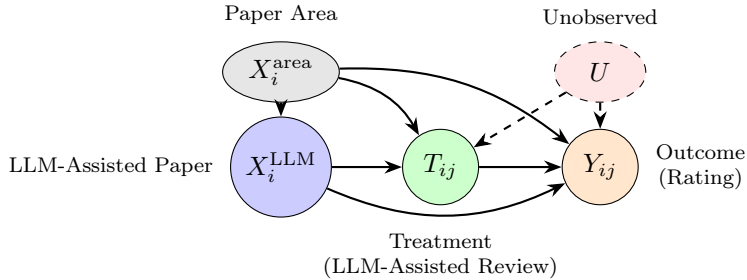

To address these concerns, we formally model this problem with the causal model in Figure \ref{fig:causal}. Solid arrows represent hypothesized causal effects; dashed arrows indicate unobserved confounding paths.
For each paper-review pair $(i,j)$, we define the treatment $T_{ij} \in \{0,1\}$ as whether review slot $j$ for paper $i$ was assigned to a reviewer that used LLM assistance, $X_i$ as paper covariates including $X_i^{\text{LLM}} \in \{0,1\}$ (whether paper $i$ contains LLM-assisted text) and $X_i^{\text{area}} \in \{0,1\}^K$ (indicator variables for which of $K$ subject areas paper $i$ belongs to), and the outcome $Y_{ij}$ as the overall rating (or auxiliary scores like presentation, contribution, or soundness).

Let $Y_{ij}(1)$ and $Y_{ij}(0)$ denote the potential outcomes under treatment and control, respectively. Our key identifying assumption is \textbf{conditional ignorability}:
\begin{equation}
T_{ij} \perp \{Y_{ij}(0), Y_{ij}(1)\} \mid X_i
\end{equation}
That is, conditioned on available covariates (in our case, the paper's area and the paper's own LLM-assistance status), treatment assignment is independent of potential outcomes. We condition on area because certain subject areas may have both higher rates of LLM-assisted reviewing and systematically different rating distributions, acting as a potential confounder.
We are specifically interested in whether the treatment effect of LLM-assisted reviewing differs depending on whether the paper itself was written with LLM assistance. Each paper has multiple review slots, which we treat as interchangeable, and receives some number of unique reviews.
As a consequence of OpenReview overwriting original ratings and making them inaccessible after the rebuttal stage, we can only observe the final ratings provided by each review.  This raises concerns about violations of the Stable Unit Treatment Value Assumption (SUTVA), which requires that (1) a unit's potential outcomes depend only on its own treatment assignment, not on the treatment of other units (no interference), and (2) there are no hidden variations in treatment (\citet{rubin}). However, several factors suggest these violations are unlikely to create stronger (hence spurious) effects than would exist in pre-rebuttal scores. First, most reviews (75-81\%) remain unchanged \citet{iclrpaper}, meaning our data predominantly reflects initial, independent assessments. Second, co-reviewer influence tends to push reviews toward consensus \citet{iclrpaper}, which would dampen rather than amplify any differential treatment patterns. While access to pre-rebuttal scores would enable stronger causal claims, these considerations suggest post-rebuttal interference is unlikely to generate the interaction effects we document.

We now frame the problem as a regression, which allows us to isolate the effects of interest while controlling for potential confounders. We use the conventional way of including interactions in a linear model by modeling the rating $Y_{ij}$ for paper $i$ and review $j$ as
\begin{equation}
Y_{ij} = \beta_0 + \beta_1 X_i^{\text{LLM}} + \beta_2 T_{ij} + \beta_3 (X_i^{\text{LLM}} \times T_{ij}) + \gamma' Z_i + \epsilon_{ij},
\label{eq:regression}
\end{equation}
where $Z_i$ is a vector of confounders we control for by including them in our regression analysis (for our regression, this is the group of indicator variables $X_i^\text{area}$ that take a value of 1 if the paper belongs to a particular area and 0 otherwise) and $\epsilon_{ij}$ is the noise term.
This framing is useful because it naturally lends significant meanings to the regression coefficients:

\begin{itemize}
    \item $\beta_0$: The expected rating for a human-written paper receiving a human-written review (the baseline).
    \item $\beta_1$: The difference in expected rating for LLM-assisted vs.\ human-written papers, holding review type constant.
    \item $\beta_2$: This is the Conditional Average Treatment Effect (CATE) of receiving an LLM review instead of a human review, for papers written without LLM assistance:
    $\text{CATE}(X_i^{\text{LLM}} = 0) = \beta_2$.
    \item Similarly, the CATE for LLM-assisted papers is given by:
    $\text{CATE}(X_i^{\text{LLM}} = 1) = \beta_2 + \beta_3$.
    This is the additional score an LLM-assisted paper receives if it is reviewed with LLM-assistance instead of entirely by a human.
\end{itemize}

The interaction coefficient $\beta_3$ is our primary quantity of interest. It captures the \textit{relative} leniency of LLM reviews toward LLM-assisted papers, compared to their leniency toward human-written papers. More precisely, it answers the question: how much more (or less) does switching from a human reviewer to a reviewer using LLM assistance benefit an LLM-assisted paper than it benefits a human-written paper?
This connects directly to our earlier summary analysis, where we observed that LLM reviews tended to assign higher ratings than human reviews to \textit{both} LLM-assisted and human-written papers on average, but the magnitude of this gap was notably larger for LLM-assisted papers (0.63 points) than for human-written papers (0.25 points). The interaction coefficient $\beta_3$ formalizes this comparison:
\begin{equation}
\beta_3 = \text{CATE}(X_i^{\text{LLM}} = 1) - \text{CATE}(X_i^{\text{LLM}} = 0)
\end{equation}
A positive $\beta_3$ would imply a positive differential kindness shown by LLM-aided reviews to LLM-aided papers as compared to human reviewers.

The results of our regression analysis on ICLR Papers and Reviews from 2024 and 2025 are detailed in Table \ref{tab:iclrfullregression}.
\begin{table}[h]
\centering
\small
\begin{adjustbox}{max width=\textwidth}  
\begin{tabular}{|l|p{7cm}|c|c|c|}
\hline
\textbf{Quantity} & \textbf{Interpretation} & \textbf{Estimate} & \textbf{95\% CI} & \textbf{p-value} \\
\hline
$\beta_0$ & Baseline: Human paper, Human review & 5.145 & [5.032, 5.259] & \textcolor{green}{<0.001} \\
\hline
$\beta_2$ & CATE: LLM-aided review effect on Human papers & +0.253 & [0.223, 0.283] & \textcolor{green}{<0.001} \\
\hline
$\beta_2 + \beta_3$ & CATE: LLM-aided review effect on LLM papers & +0.629 & [0.476, 0.782] & \textcolor{green}{<0.001} \\
\hline
$\beta_3$ & Differential: Extra kindness to LLM-aided papers & +0.376 & [0.221, 0.532] & \textcolor{green}{<0.001} \\
\hline
\end{tabular}
\end{adjustbox}
\caption{Regression Analysis for determining LLM Review Kindness using the model from Eq. (\ref{eq:regression}) on ICLR 2024+ICLR 2025 combined.}
\label{tab:iclrfullregression}
\end{table}

These results imply that LLM-aided reviewers are kinder than human reviewers to both human papers and LLM-aided papers, but their relative kindness, determined by $\beta_3$, is higher for LLM-aided papers than for human papers. This is very close to the observation we made in Table \ref{tab:iclrfullsummary}, where the differential kindness of LLM-aided reviews to LLM-assisted papers was $0.63-0.25=0.38$.
Although regression controls for observable confounders, unobserved paper-level characteristics (e.g., topic trendiness, writing clarity) may still bias our estimates if they jointly influence both the likelihood of receiving an LLM review and the rating itself. Adding additional terms to the regression formulation would eventually run into the curse of dimensionality, particularly if interactions between those terms matter. We also run into the risk of multicollinearity issues or accidentally conditioning on colliders and creating spurious associations. To address this limitation, we turn to a more stringent identification strategy that eliminates between-paper confounding without the typical drawbacks associated with potentially overspecified regressions. 

\subsubsection{Within Paper Paired Analysis}
The regression framework, despite controlling for area, still compares ratings across different papers. We now leverage a unique feature of peer review: papers receive multiple reviews, allowing us to observe both potential outcomes for the same paper. By restricting to papers that received at least one human review and at least one LLM-assisted review, we can compute the actual treatment effect for each paper without relying on conditional independence assumptions. While restricting our corpus to only papers that received both types of reviews is a limitation that can introduce a selection bias, it provides an alternative method for estimating the CATE that can corroborate the previous analysis. The restricted corpus consists of 10945 papers.

For each paper $i$, we compute the within-paper difference:
\begin{equation}
\Delta_i = \bar{Y}_i^{\text{LLM}} - \bar{Y}_i^{\text{Human}}
\end{equation}
where $\bar{Y}_i^{\text{LLM}}$ and $\bar{Y}_i^{\text{Human}}$ are the average ratings from LLM-assisted and human reviews, respectively. $\Delta_i$ is the paper-specific treatment effect, holding all paper characteristics constant.
We partition papers by LLM status and estimate:
\begin{align}
\text{CATE}(X_i^{\text{LLM}} = 0) &= \mathbb{E}[\Delta_i \mid \text{paper is human}],\qquad
\text{CATE}(X_i^{\text{LLM}} = 1) = \mathbb{E}[\Delta_i \mid \text{paper is LLM-assisted}]
\end{align}

If LLM reviews exhibit differential leniency toward LLM-assisted papers, then CATE$(X_i^{\text{LLM}} = 1) >$ CATE$(X_i^{\text{LLM}} = 0)$. Beyond comparing means, we test whether the entire distribution of treatment effects differs between paper types using the Mann-Whitney U test:
\begin{align}
H_0&: \text{The distributions of } \Delta_i \text{ for human and LLM-assisted papers are identical} \\
H_1&: \Delta_i \text{ for LLM-assisted papers is different from } \Delta_i \text{ for human papers}
\end{align}

Table \ref{tab:iclrfullmannwhitneyutest} shows the results. The mean treatment effect is 0.340 for LLM-assisted papers versus 0.244 for human papers (difference: +0.096), though the p-value of 0.058 provides only weak evidence against the null hypothesis. Comparing this number to $\beta_3$ (0.376) from Table \ref{tab:iclrfullregression} and the estimate from Table \ref{tab:iclrfullsummary} (0.38), we find the effect size to be much smaller upon controlling for paper-level confounding factors. 

\begin{table}[h]
\centering
\small
\begin{adjustbox}{max width=\textwidth}  
\begin{tabular}{|l|c|c|c|c|}
\hline
\textbf{Metric} & \textbf{CATE (LLM-aided papers)} & \textbf{CATE (Human papers)} & \textbf{Difference} & \textbf{p-value} \\
\hline
Rating & 0.340 & 0.244 & +0.096 & 0.058 \\
\hline
\end{tabular}
\end{adjustbox}
\caption{Within-paper treatment effects (ICLR 2024-2025). Mann-Whitney U test (one-tailed) for distributional difference.}
\label{tab:iclrfullmannwhitneyutest}
\end{table}

\subsubsection{How influential are LLM-aided reviews in final decisions?}

Beyond examining rating patterns, we investigate whether LLM-aided reviews carry comparable weight to human reviews in determining final acceptance decisions. We measure review influence by alignment with the final decision: a review "aligns" if its rating matches the paper's outcome (rating $> 5$ for accepted papers, rating $\leq 5$ for rejected papers). We additionally stratify by paper quality to examine whether LLM-aided reviews show differential alignment for borderline versus clear-cut cases, as influence at decision boundaries is most consequential for outcomes.

Using the ICLR 2024 and 2025 combined corpus, we stratify papers by quality using a hold-out human review as the quality indicator, then compute alignment rates for LLM-aided and human reviews (excluding the quality-indicator review) across six quality buckets that correspond to the six ratings reviewers are able to choose from. Table \ref{tab:review_alignment} shows the results. Human reviews consistently show higher alignment rates than LLM-aided reviews across most quality levels, with alignment scores between $75\%-87\%$ compared to $69\%-81\%$ for LLM-aided reviews.  

\begin{table}[h]
\centering
\begin{tabular}{|c|c|c|c|c|}
\hline
\textbf{Paper Quality} & \textbf{Human Align \%} & \textbf{LLM Align \%} \\
\hline
1 & 87.4 & 80.7 \\
3 & 80.8 & 73.5  \\
5 & 75.0 & 69.5 \\
6 & 76.0 & 74.9 \\
8 & 78.8 & 79.2 \\
10 & 79.9 & 74.6  \\
Overall & 77.7 & 73.8 \\
\hline
\end{tabular}
\caption{Review alignment rates with final decisions by paper quality for ICLR 2024-2025. Alignment defined as: rating $> 5$ for accepted papers, rating $\leq 5$ for rejected papers. Quality determined by hold-out human review.}
\label{tab:review_alignment}
\end{table}

These patterns suggest human reviews remain more influential in decision-making, even for borderline papers where the accept/reject decision is most consequential. The lower alignment of LLM-aided reviews in the medium-quality range (ratings 3-6) can be an artifact of their tendency to inflate ratings for lower quality papers relative to human reviews, which leads to misalignment in rejection decisions. In fact, $91.6\%$ of LLM-aided review misalignments in quality bucket 5 are due to recommended acceptances for ultimately rejected papers.

At this stage, the results indicate evidence for LLM-aided reviews favoring LLM-aided papers, though their lower alignment rates seem to suggest limited influence on final decisions. We now turn to accepted papers to see if this pattern persists, since these papers form the most important paper subset in the corpus.

\subsection{Does the kindness of LLM-aided reviews to LLM-aided papers vary with paper quality?}
Interestingly, when we restrict the ICLR corpus to only accepted papers, we observe markedly different patterns (Table \ref{tab:iclracceptedsummary}).
\begin{table}[b]
\centering
\small
\begin{adjustbox}{max width=\textwidth}  
\begin{tabular}{|l|c|c|c|c|c|c|}
\hline
\textbf{Quadrant }\textit{(Paper, Review)} & \textbf{Rating} & \textbf{Confidence} & \textbf{Presentation} & \textbf{Contribution} & \textbf{Soundness} \\
\hline
(Human, Human) & $6.467 \pm 1.319$ & $3.573 \pm 0.792$ & $2.902 \pm 0.681$ & $2.720 \pm 0.632$ & $2.918 \pm 0.605$ \\
\hline
(Human, LLM) & $6.409 \pm 1.226$ & $3.535 \pm 0.811$ & $2.932 \pm 0.604$ & $2.756 \pm 0.621$ & $2.924 \pm 0.555$ \\
\hline
(LLM, Human) & $6.482 \pm 1.309$ & $3.671 \pm 0.752$ & $2.879 \pm 0.752$ & $2.725 \pm 0.610$ & $2.839 \pm 0.598$ \\
\hline
(LLM, LLM) & $6.552 \pm 1.242$ & $3.606 \pm 0.853$ & $2.945 \pm 0.566$ & $2.836 \pm 0.656$ & $2.903 \pm 0.597$ \\
\hline
\end{tabular}
\end{adjustbox}
\caption{Summary Statistics for (Paper, Review) pairs from ICLR 2024 and 2025 combined (accepted papers only). (Paper, Review) pairs are divided into 4 quadrants based on the LLM status of the paper and the review. $\alpha$ value of 0.15 is used as a threshold: Papers and Reviews are classified as LLM-aided if they have $\alpha>0.15$ and Human otherwise. Cell values are presented as Mean $\pm$ Std Dev}
\label{tab:iclracceptedsummary}
\end{table}
The quadrants now show similar scores with no apparent differential LLM kindness. We observe similar patterns for NeurIPS and ICML (Appendix \ref{sec:appsummary}), where by default rejected papers are private on OpenReview (though some authors choose to make their papers public).
\subsubsection{Regression with area correction}
In order to investigate this, we run the same regression analysis from Section \ref{sec:reg} on the accepted only corpus of ICLR 2024-2025,  in Table \ref{tab:iclr_accepted_regression}. 
The previously observed differential kindness ($\beta_3$) is no longer significant (p > 0.30), and individual CATEs are near zero. Similar null results appear for NeurIPS and ICML (Appendix \ref{sec:appregneu}, \ref{sec:appregicml}).

\begin{table}[t]
\centering
\small
\begin{adjustbox}{max width=\textwidth}  
\begin{tabular}{|l|p{7cm}|c|c|c|}
\hline
\textbf{Quantity} & \textbf{Interpretation} & \textbf{Estimate} & \textbf{95\% CI} & \textbf{p-value} \\
\hline
$\beta_0$ & Baseline: Human paper, Human review & 6.544 & [6.389, 6.699] & \textcolor{green}{<0.001} \\
\hline
$\beta_2$ & CATE: LLM-aided review effect on Human papers & $-0.051$ & [$-0.091$, $-0.012$] & 0.011 \\
\hline
$\beta_2 + \beta_3$ & CATE: LLM-aided review effect on LLM-aided papers & $+0.082$ & [$-0.168$, $+0.332$] & 0.520 \\
\hline
$\beta_3$ & Differential: Extra kindness to LLM-aided papers & $+0.134$ & [$-0.120$, $+0.387$] & 0.301 \\
\hline
\end{tabular}
\end{adjustbox}
\caption{Regression Analysis for ICLR accepted papers only (2024+2025 Combined). Model: Rating = $\beta_0 + \beta_1 X_i^{\text{LLM}} + \beta_2 T_{ij} + \beta_3 (X_i^{\text{LLM}} \times T_{ij}) + \gamma' Z_i$. Area fixed effects included. The interaction effect $\beta_3$ is not statistically significant, in contrast to the full dataset (Table~\ref{tab:iclrfullregression}).}
\label{tab:iclr_accepted_regression}
\end{table}

\begin{figure}[h]
\centering
\begin{minipage}{0.3\textwidth}
    \centering
    \includegraphics[width=\textwidth]{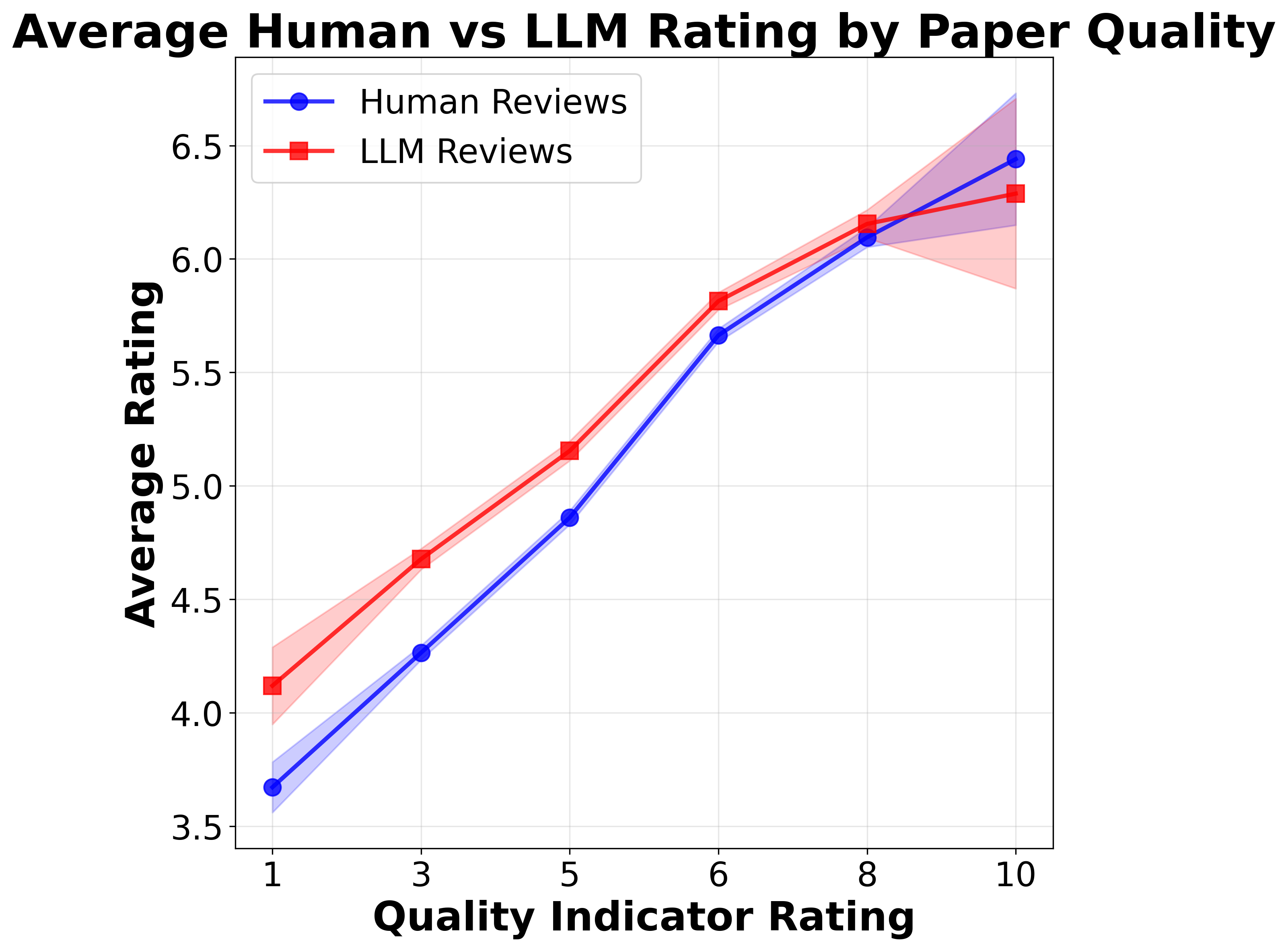}
    \caption{Average ratings from LLM-aided reviews vs. human reviews across quality buckets. Shaded area is 1.96*std-error.}
    \label{fig:llm_leniency_by_quality}
\end{minipage}
\hfill
\begin{minipage}{0.68\textwidth}
    \centering
    \includegraphics[width=\textwidth]{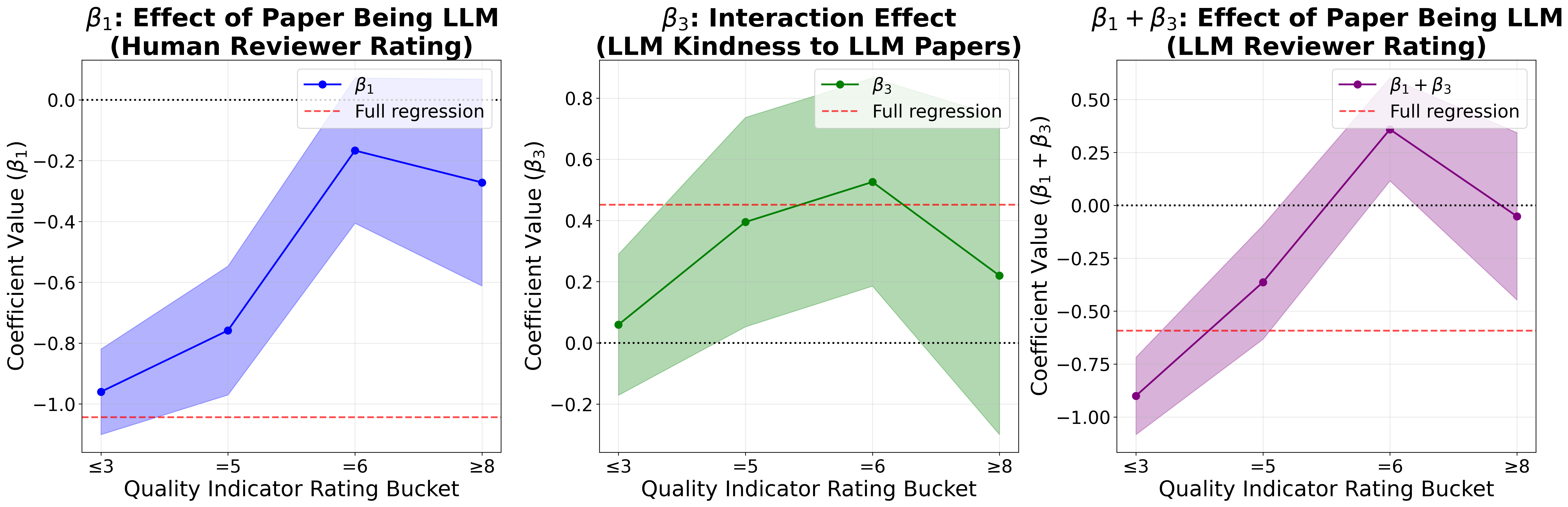}
    \caption{Regression coefficients by quality bucket. Each point represents a separate regression run on papers within that quality bucket. Shaded area is 95\% CI. The dashed red line shows the estimate from the full regression for comparison. Zero is indicated with a black dotted line.}
    \label{fig:quality_buckets}
\end{minipage}
\end{figure}

To understand this discrepancy between results on the accepted-only corpus and the all papers corpus, we examine whether LLM-aided reviews differ from human reviews across the quality spectrum. We stratify papers into four quality buckets (rating $\leq 3$, $= 5$, $= 6$, $\geq 8$) using a randomly selected held-out human review as the quality indicator, then compute average ratings from LLM-assisted and human reviews (excluding the quality-indicator review) within each bucket.
Figure \ref{fig:llm_leniency_by_quality} reveals two key patterns. First, LLM-aided reviews provide substantially higher ratings than human reviews for low-quality papers, but this gap narrows and disappears as paper quality increases. For papers with quality indicator rating $\leq 3$, LLM-aided reviews average approximately 0.4-0.5 points higher than human reviews, shrinking to near-zero for medium and high-quality papers. Second, LLM-assisted papers are heavily overrepresented in the low-quality bucket: 173 of 2,768 papers (6.2\%) in the rating $\leq 3$ bucket are LLM-assisted, compared to only 29 of 1,316 papers (2.2\%) in the rating $\geq 8$ bucket—mirroring the rating distribution patterns in Figure \ref{fig:paperratingdist}. The combination of these two factors could potentially be creating the appearance of differential kindness in aggregate analyses. 

We probe this quality-dependent relationship further by running separate regressions within each quality bucket using all remaining reviews (excluding the quality-indicator review). Figure \ref{fig:quality_buckets} plots the regression coefficients for each bucket. The leftmost panel shows that $\beta_1$ (the effect of a paper being LLM-assisted on ratings from human reviewers) is strongly negative for low-quality papers ($\approx -1.0$ for rating $\leq 3$) but approaches zero for high-quality papers (rating $\geq 8$)—human reviewers penalize low-quality LLM-aided papers but show no bias for high-quality LLM-aided papers. The center panel shows the interaction effect $\beta_3$ follows an inverted-U pattern: it is small for the lowest-quality papers ($\approx 0.1$), peaks around medium quality (rating 5-6, where $\beta_3 \approx 0.5$), and decreases again for high-quality papers ($\approx 0.2$).  Notably, $\beta_3$ remains positive across all buckets but only achieves statistical significance in the middle-quality range, where the regression estimate on the full population (dashed red line) aligns with the bucket-specific estimates. This suggest LLM-aided reviews show differential kindness to LLM-aided papers in the intermediate quality range, but not otherwise. The rightmost panel shows $\beta_1 + \beta_3$ (the effect of a paper being LLM-aided on ratings from LLM-aided reviewers) is strongly negative for low-quality papers but becomes slightly positive for medium-quality papers before returning to near-zero for high-quality papers - LLM-aided reviewers penalize low-quality LLM-aided papers (but to a lesser extent than human reviewers) and only show a very small bias for high-quality LLM-aided papers. 

These patterns reconcile our earlier findings: the positive $\beta_3$ estimated by regression over the complete paper corpus reflects the middle-quality range where the effect is strongest. Restricting to the accepted-papers corpus is similar to restricting to the high-quality range where $\beta_3$ approaches zero.

\subsubsection{Within Paper Paired Analysis.}
To further validate these findings, we repeat the within-paper paired analysis stratified by quality bucket. For each bucket, we compute treatment effects $\Delta_i$ using only the non-quality-indicator reviews and test whether $\Delta_i$ differs between LLM and human papers.
Table \ref{tab:quality_bucket_paired} shows the results. Interestingly, none of the quality buckets show statistically significant differences in treatment effects between LLM and human papers, though point estimates vary.
We observe that the effect magnitudes are also slightly smaller than for the entire paper corpus (0.096 points), and even change sign in the `Med-High' bucket, except for papers in the `High Quality' bucket, which have a low frequency of LLM-aided papers.
\begin{table}[h]
\centering
\begin{tabular}{|l|c|c|c|c|c|}
\hline
\textbf{Quality Bucket} & \textbf{N$_{\text{LLM}}$} & \textbf{N$_{\text{Human}}$} & \textbf{CATE Diff.} & \textbf{p-value} & \textbf{Sig.} \\
\hline
Low (rating $\leq 3$) & 173 & 2,595 & +0.038 & 0.332 & no \\
Medium (rating $= 5$) & 66 & 2,723 & +0.082 & 0.292 & no \\
Med-High (rating $= 6$) & 65 & 2,831 & $-0.090$ & 0.524 & no \\
High (rating $\geq 8$) & 29 & 1,287 & +0.369 & 0.137 & no \\
\hline
\end{tabular}
\caption{Within-paper paired analysis by quality bucket. CATE Diff. = mean($\Delta_i$ for LLM papers) $-$ mean($\Delta_i$ for human papers). N$_{\text{Human}}$ and N$_{\text{LLM}}$ represent the number of Human and LLM-aided papers in the bucket. Mann-Whitney U test (one-tailed).}
\label{tab:quality_bucket_paired}
\end{table}

The lack of significance within buckets, despite regression showing varying $\beta_3$, suggests aggregate patterns reflect between-paper heterogeneity rather than systematic differential treatment. Combined with LLM-aided reviews' uniform leniency toward low-quality papers (Figure \ref{fig:llm_leniency_by_quality}), this supports the interpretation that apparent LLM-LLM interaction is driven by LLM-aided papers' overrepresentation in low-quality ranges where LLM-aided reviews are generally lenient, not by specific favoritism toward LLM-aided content.

\subsection{How do fully LLM-generated reviews differ from LLM-aided reviews?}
Up to this point, our analysis has relied on observational data from real papers and reviews, capturing authentic peer review dynamics. However, this conflates two distinct agents: the LLM and the human using it. When we classify a review as \textit{LLM-assisted}, we identify reviews with LLM-aided text, but the final rating reflects a human reviewer who may accept, modify, or ignore any LLM suggestions. Our observational analysis captures \textit{humans who use LLMs}, not LLMs themselves. This is an important distinction since identical prompts can produce markedly different responses across sessions and user contexts \citet{wang2025inadequacyofflinellmevaluations}. This raises the following question: if we see some form of bias in LLM-assisted reviews, is this bias due to the LLM or due to the human using the LLM?

To disentangle these effects, we augment our analysis with experiments on fully LLM-generated reviews. 
In particular, we prompt GPT-4o to generate reviews for papers in our corpus, obtaining ratings reflecting the LLM's judgment directly, without human mediation. 
We avoid GPT-5 because of the strong possibility that it has been trained on the original reviews already. 
This isolates the \textit{pure LLM} effect, allowing comparison to both human reviewers and \textit{humans using LLMs}. Note that the fully LLM-generated reviews are not subject to confounding or selection biases, since they are synthetically generated.

We generate reviews for randomly sampled papers from the 2025 editions of ICLR, ICML, and NeurIPS. Prompts follow the template used by \citet{lu2024aiscientistfullyautomated} with conference-specific guidelines and no additional instructions to be lenient or harsh (details in Appendix \ref{sec:appprompts}).
We sample 60 LLM-aided papers and 480 human papers from the ICLR 2025 corpus. Furthermore, we sample 75 LLM-aided papers and 300 human papers from the ICLR 2025 accepted corpus, the NeurIPS 2025 corpus, and the ICLR 2025 corpus. Note that this oversamples LLM-aided papers to ensure reliable statistical analysis. These numbers are restricted by the number of LLM-aided papers available in the corpus, since ICML 2025 only has 75 papers classified as LLM-aided.
We then perform the same within-paper paired analysis on these generated reviews against the human reviews for the same papers. Table \ref{tab:generatedresults} has these results for the "Overall Rating" field while additional results are included in Appendix \ref{sec:appwithin}.
\begin{table}[h]
\centering
\small\begin{adjustbox}{max width=\textwidth}  
\begin{tabular}{|l|l|c|c|c|c|c|c|}
\hline
\textbf{Conference} & \textbf{Range} & \textbf{CATE (LLM Papers)} & \textbf{CATE (Human Papers)} & \textbf{Difference} & \textbf{p-value} \\
\hline
ICLR 2025 & 1-10 & 2.248 & 1.623 & +0.625 & 0.0018 \\
\hline
ICLR 2025 (Accepted) & 1-10 & 0.674 & 0.712 & $-0.038$ & 0.5670 \\
\hline
NeurIPS 2025 & 1-6 & 0.408 & 0.383 & +0.025 & 0.3616  \\
\hline
ICML 2025 & 1-5 & 0.774 & 0.759 & +0.015 & 0.3315 \\
\hline
\end{tabular}
\end{adjustbox}
\caption{Result of Mann-Whitney U test (one-tailed) on 2025 conference data with fully generated reviews. Kindness difference = Generated Review rating - Human rating.}
\label{tab:generatedresults}
\end{table}
These results are interesting in two respects. 
First, we notice that the magnitude of difference in mean kindness for LLMs over Humans on LLM Papers as compared to Human Papers is actually a lot larger than what we saw with observational data in \ref{tab:iclrfullmannwhitneyutest} where it was only about 0.1. This could be explained by a few hypotheses: for example, people using LLMs might either lower the rating the LLM provides, prompt the LLM to directly provide a more critical review, or merely use the LLM to expand upon their own critical review.
Second, the generated reviews provide substantially higher ratings than human reviews for both human and LLM-aided papers but again show no differential treatment when we restrict the corpus to only accepted papers.

\begin{figure}[h]
\centering
\begin{minipage}{0.45\textwidth}
    \centering
    \includegraphics[width=\textwidth]{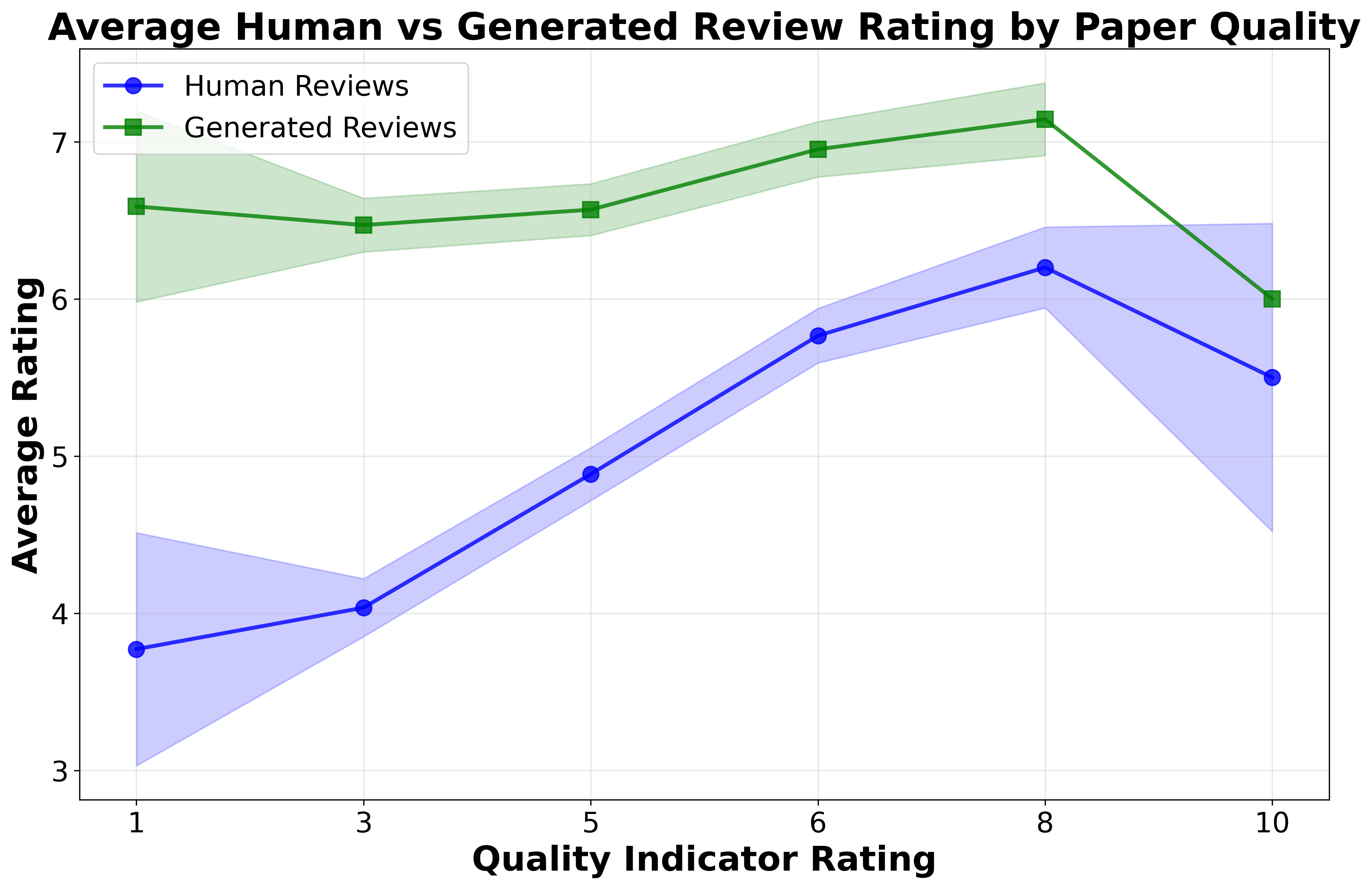}
    \caption{Average Rating provided by Human reviews and fully LLM reviews to randomly sampled ICLR 2025 papers, bucketed by quality. The datapoints correspond to quality bucket 10 are noisy because only 1 such paper was sampled.}
    \label{fig:generatelines}
\end{minipage}
\hfill
\begin{minipage}{0.45\textwidth}
    \centering
    \includegraphics[width=\textwidth]{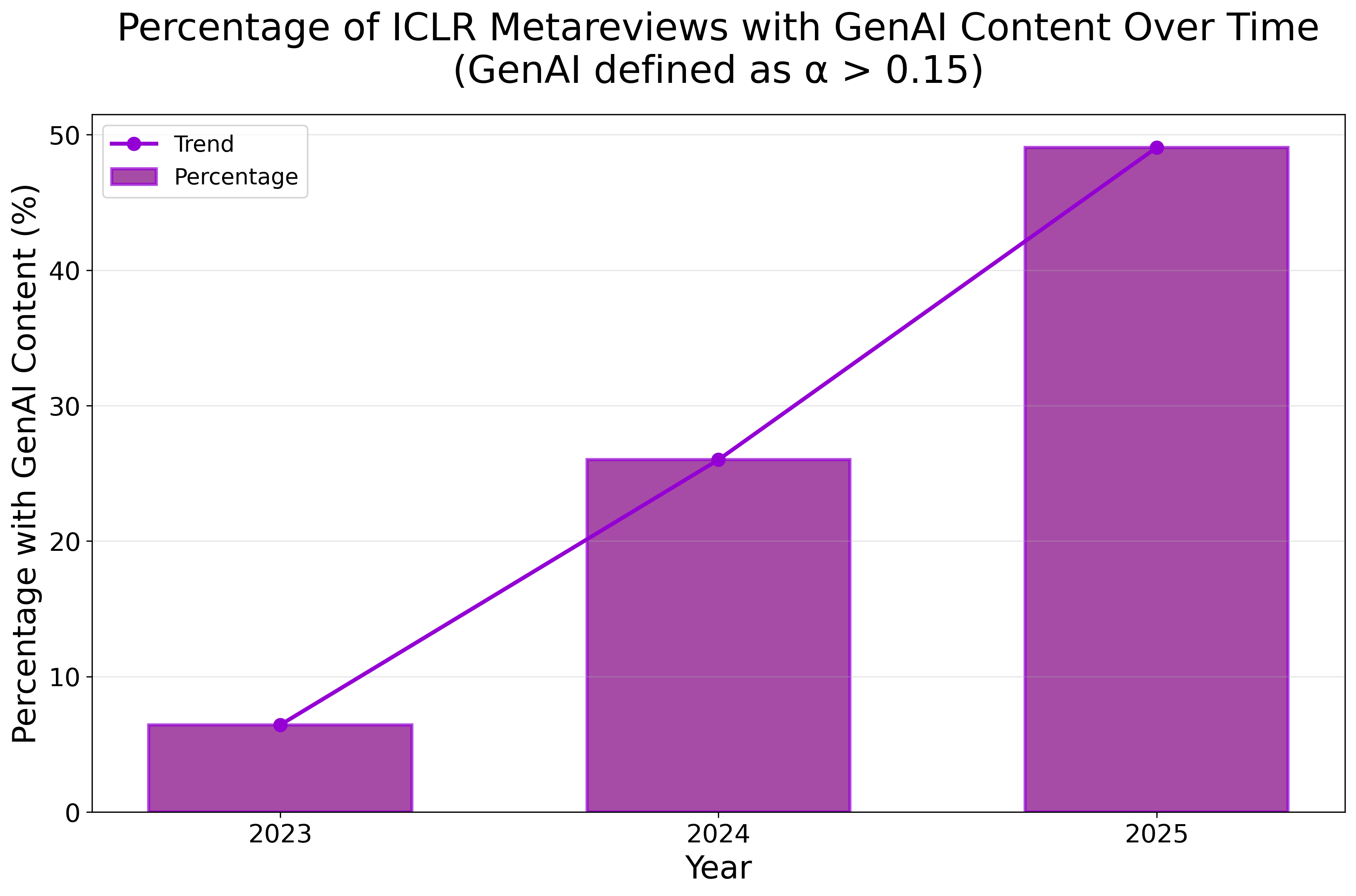}
    \caption{Occurence of LLM-aided metareviews in ICLR. An LLM-aided metareview has $\alpha>0.15$. We see an increasing trend with over $49\%$ of metareviews in 2025 satsisfying this criterion.}
    \label{fig:metareviewalphatrend}
\end{minipage}
\end{figure}

We next examine the distribution of scores assigned by these fully LLM-generated reviews (Figure~\ref{fig:generatelines}). We observe that fully LLM-generated reviews assign ratings almost exclusively between 6 and 7 regardless of paper quality, exhibiting minimal slope across quality buckets. This compression has serious implications: fully LLM-generated reviews fail to discriminate paper quality, systematically inflating ratings for weak papers while providing little signal about paper merit. The gap between LLM and human ratings is largest for low-quality papers, where automated reviews would recommend acceptance for papers humans would reject. These observations suggest that fully automated LLM reviewing would undermine the accuracy of peer review. Human-in-the-loop moderation (as seen in observational LLM-assisted reviews) partially mitigates this compression: the line corresponding to LLM-aided reviews in Figure \ref{fig:llm_leniency_by_quality} has a lower slope and higher intercept than the line corresponding to human reviews, but still shows some discriminative power. 

The results of the significance test indicate that fully LLM-generated reviews are not especially kinder to LLM-aided papers than they are to Human papers. Instead, they are simply less harsh to papers in general, and the over-representation of LLM aided papers in low quality buckets only misleads us into thinking this differential kindness exists in the complete ICLR corpus.

\subsection{How does LLM-assistance in metareviewing relate to a paper's final acceptance decision?}
\label{sec:metareview}
Metareviews are written by Area Chairs based on the reviews to render a final acceptance decision. While metareviewers can add their own analysis and arguments, part of their role is summarizing the arguments provided by the reviewers. Summarization is a domain where LLMs typically excel, making metareviews a tempting target for LLM assistance. 

Applying the detection method from \citet{liaodetection} to ICLR metareviews from 2023-2025, we observe increasing LLM use (Figure~\ref{fig:metareviewalphatrend}), with ICLR 2025 showing evidence of significant LLM involvement in up to $49\%$ of metareviews. This indicates a growing rate of LLM adoption even among senior researchers in the field.

Based on the role of a metareview in the peer review process, a key question is whether an LLM-assisted metareview is more or less likely to render an accept decision than a human metareview observing the same set of scores. 
\subsubsection{Regression Analysis}
We frame this as a logistic regression problem for every individual paper:
\begin{equation}
\text{logit}(P_{\text{accepted}}) = \beta_1(\text{mean rating}) + \beta_2(\text{std rating}) + \beta_3I_{\text{metareview}} + \gamma' Z,
\label{eq:logistic_acceptance}
\end{equation}
where $P_{\text{accepted}}$ is the probability of paper acceptance, mean rating captures the central tendency of reviewer scores, std rating measures the disagreement among reviewers, $I_{\text{metareview}}$ is a binary indicator for LLM-aided metareviews ($\alpha > 0.15$), and $Z$ represents area fixed effects.
Our coefficient of interest, $\beta_3$, measures whether LLM-aided metareviews associate with different acceptance probabilities after controlling for both rating level, agreement, and area.

Table~\ref{tab:iclr20245_logistic_acceptance} presents the results for ICLR 2024 and 2025 combined. As expected, higher mean ratings strongly predict acceptance ($\beta_1 = +3.327$, $p < 0.001$, 95\% CI: [3.323, 3.331]), while higher rating variance reduces acceptance likelihood ($\beta_2 = -0.288$, $p < 0.001$, 95\% CI: [-0.297, -0.280]).
Importantly, $\beta_3 = +0.214$ ($p < 0.001$, 95\% CI: [+0.204, +0.224]), indicating that papers with LLM-aided metareviews are significantly more likely to be accepted, after controlling for both the mean and variance of ratings. This suggests that LLM assistance in metareview writing is associated with more favorable acceptance decisions beyond what would be expected from the underlying review scores alone.

\begin{table}[h]
\centering
\small
\begin{adjustbox}{max width=\textwidth}  
\begin{tabular}{|l|p{6cm}|c|c|c|}
\hline
\textbf{Parameter} & \textbf{Interpretation} & \textbf{Coefficient} & \textbf{95\% CI} & \textbf{p-value} \\
\hline
$\beta_1$ & Contribution of mean rating & $+3.327$ & [3.323, 3.331] & \textcolor{green}{$<0.001$} \\
\hline
$\beta_2$ & Contribution of std deviation of ratings & $-0.288$ & [-0.297, -0.280] & \textcolor{green}{$<0.001$} \\
\hline
$\beta_3$ & Contribution of metareview being LLM-assisted & $+0.214$ & [+0.204, +0.224] & \textcolor{green}{$<0.001$} \\
\hline
\end{tabular}
\end{adjustbox}
\caption{Logistic regression results for ICLR 2024-2025. Model: Eq.~\ref{eq:logistic_acceptance}. Target: Accept (1) vs Reject (0). Area fixed effects included. Higher mean ratings increase acceptance probability, higher rating variance decreases it, and LLM-aided metareviews are associated with increased acceptance likelihood controlling for both rating level and consensus.}
\label{tab:iclr20245_logistic_acceptance}
\end{table}

\subsubsection{Fully LLM-generated Metareviews}
To again disentangle whether the decisions that humans make with LLM-assistance align with the decisions that LLMs make on their own, we generated fully LLM-generated metareviews for 300 randomly sampled ICLR 2025 papers. 70 of these papers do not have metareviews because they were withdrawn from the conference after initial reviews, which raises the acceptance rate in the remaining corpus of 230 papers to 46.5\%. We prompt GPT-4o with all reviews a paper received and the conference rating and confidence scales (see Appendix \ref{sec:appprompts} for full prompts). We then extract accept/reject decisions from the generated metareviews and compare these fully LLM-generated decisions to actual decisions, stratified by whether the actual metareview was LLM-aided ($\alpha > 0.15$).

Table \ref{tab:synthetic_metareview_alignment} shows the results. Overall, fully LLM-generated metareviews align with actual decisions 72.6\% of the time. We actually see that fully LLM-generated metareviews are substantially harsher than actual decisions: they recommend acceptance for only 44 of 230 papers (19.1\%), while actual decisions accepted 107 papers (46.5\%). Alignment rates differ by actual metareview type: synthetic decisions align with human metareviews 76.9\% of the time versus 68.9\% for LLM-aided metareviews (which tended to be more positive).

\begin{table}[t]
\centering
\small
\begin{adjustbox}{max width=\textwidth}  
\begin{tabular}{|l|c|c|c|c|}
\hline
\textbf{Actual Metareview Type} & \textbf{N} & \textbf{Alignment \%} & \textbf{Synthetic Accept Rate} & \textbf{Actual Accept Rate} \\
\hline
Human ($\alpha \leq 0.15$) & 108 & 76.9 & 17.6\% & 40.7\% \\
LLM-aided ($\alpha > 0.15$) & 122 & 68.9 & 20.5\% & 51.6\% \\
\hline
\textbf{Overall} & 230 & 72.6 & 19.1\% & 46.5\% \\
\hline
\end{tabular}
\end{adjustbox}
\caption{Alignment between fully LLM-generated metareviews and actual decisions (ICLR 2025 sample). Synthetic metareviews are substantially harsher than actual decisions, recommending acceptance at less than half the rate.}
\label{tab:synthetic_metareview_alignment}
\end{table}

For both human and LLM-aided metareviews, fully LLM-generated decisions achieve 100\% precision (when they recommend accept, the paper was actually accepted) but low recall (43.2\% for human metareviews, 39.7\% for LLM-aided). Fully LLM-generated metareviews systematically under-predict acceptances: they correctly identify strong accepts but fail to recommend acceptance for many borderline papers that human metareviewers ultimately accepted. While we can only speculate about how metareviewers use the LLM as part of their decision process, they do not appear to be following the LLM acceptance recommendations without further reflection.

The harshness of fully LLM metareviews also reveals how LLMs interpret reviewing instructions. Despite receiving the same reviews and guidelines that led human metareviewers to accept 40.7\% of papers, the LLM recommends acceptance for only 17.6\%. This suggests LLMs may have stricter interpretations of decision thresholds, or a lack of context typically associated with scores, for instance, a rating of '8' is only termed an "Accept" but is generally considered a very good paper. This would also explain their tendency to give out higher ratings as reviewers when recommending acceptances, even when following identical instructions.

\section{Key Takeaways}
We return to the research questions introduced in Section \ref{sec:intro} with empirically-grounded answers:

\textit{Are LLM-aided reviews kinder to LLM-aided papers on average and does this vary by paper quality?}
Our threshold-varying analysis (Figure \ref{fig:quality_buckets}) demonstrates that the interaction effect is strongly modulated by paper quality. For papers in the top quality buckets, we observe minimal differential treatment ($\beta_3 \approx 0$). The positive interaction coefficient observed in the full corpus is driven primarily by papers in the lower quality buckets, where LLM reviews exhibit greater leniency than human reviews, a leniency that benefits LLM-assisted papers more simply because they are overrepresented in this range. As shown in Figure \ref{fig:llm_leniency_by_quality}, LLM-assisted reviews provide systematically higher ratings to lower quality papers regardless of whether those papers used LLM assistance, but this compression of the rating scale disproportionately affects the (LLM Paper, LLM Review) quadrant due to its compositional skew toward weaker submissions. The within-paper paired analyses with quality buckets (Table \ref{tab:quality_bucket_paired}) confirms this observation.

\textit{How does an LLM-aided review differ from a fully LLM-generated review? What is the effect of the human in the loop?}
Our synthetic experiments (Table \ref{tab:generatedresults}) reveal substantial differences between unmediated LLM reviews and human-mediated LLM-assisted reviews. Fully LLM-generated reviews exhibit markedly compressed rating distributions, assigning scores almost exclusively in the 6–7 range (on a 1–10 scale) regardless of paper quality (Figure \ref{fig:generatelines}). LLM-aided reviews have a broader and more discriminating rating distribution that better tracks paper quality. The kindness differential between LLM and human reviews for LLM-assisted papers is approximately 0.625 points in the fully synthetic condition (ICLR 2025 full corpus) but only 0.096 points in the observational data (Table \ref{tab:iclrfullmannwhitneyutest}), a reduction of roughly 85\%.

\textit{How does LLM use in a metareview relate to a paper's final decision?}
The regression analyses in Section \ref{sec:metareview} reveals that given the same set of scores, an LLM-aided metareviewer is more likely to render an accept decision than a human reviewer. However, synthetic experiments show that fully LLM metareviews are actually significantly harsher than their human and LLM-aided counterparts.
\section{Discussion and Conclusions}
This paper provides empirical insights into how LLM use on both sides of peer review affects ratings and paper outcomes. We find asymmetric patterns: while LLM-aided papers receive lower ratings overall, LLM-aided reviews show stronger positive effects when evaluating LLM-aided papers. Critically, these patterns largely disappear when restricting analysis to accepted papers, indicating effects are driven by lower quality submissions where LLM-aided reviews exhibit general leniency. The degree of rating compression, both in LLM-aided and fully LLM-generated reviews (the latter corroborating \citet{zhu2025reviewerllmbiasesdivergence}'s findings) is particularly concerning since these reviews ultimately provide less information and introduce noise into the decision process. Beyond review ratings, we find that LLM-aided metareviews are associated with higher acceptance rates despite fully LLM-generated metareviews being overly conservative, and that LLM-aided reviews align less well with final decisions than human reviews, even for borderline papers where decision-making is most consequential. We also make observations about how LLM-aided reviews differ from their fully LLM-generated counterparts, noting the latter's lack of discriminative power. 

Periodic reviews of this kind are essential as this landscape continues to evolve. Interesting future directions include comparing our results to those from submissions to similarly reputed journals, which are often less susceptible to some of the issues conferences face: reviewers are not authors competing for the same limited spots, and review timelines are less stressed. We also aim to examine how this relationship varies across subjects, not limited to computer science. However, conferences in other domains (and even journals within CS) do not provide numerical scores with their reviews, which is a major component of our present analysis.

Our results on LLM-aided and fully LLM-generated reviews/metareviews together suggest potential patterns in the way people use LLMs in peer review. For instance, metareviewers might selectively deploy LLMs when writing accept decisions, where the summarization task is more straightforward, but write rejection metareviews themselves, where detailed critical reasoning may be required. Alternatively, LLM use could be more popular among metareviewers who are more lenient. Similarly, the gap between LLM-aided reviews and fully LLM-generated reviews could be a result of reviewers asking LLMs to be more critical in their prompts, or they could simply be using LLMs only to expand upon their own salient reviews. We encourage full-scale user studies to test any such hypotheses.

Multiple major ML conferences have now instituted LLM policies beyond simple disclosure of use. On the most permissive end, the Association for the Advancement of Artifical Intelligence started a pilot program (\citet{aaai2026peerreview}) to provide supplementary LLM-generated reviews to authors alongside human reviews, and provide LLM generated summaries of author-reviewer discussions to aid metareviewers in their decision. As mentioned in Section \ref{sec:intro}, ICML has instituted a new policy for their 2026 edition requiring both reviewers and authors to declare either conservative or permissive LLM use, with reviewer-author pairing done based on these declarations. We see this as problematic, since our results indicate that this could lead to a larger number of low-quality LLM-aided papers being accepted, because of the general leniency of LLM-assisted and fully LLM generated reviews. The academic community needs to direct their focus to identifying ways to use these tools in a way that is beneficial to the peer review process, and to counteract use that is detrimental. We hope the findings of this paper provide valuable insights to conference organizers as they design new policies and tools.  

\section{Generative AI usage statement}
Generative AI tools have been used to assist in formatting \LaTeX (creating tables and improving alignment). 

\paragraph{Acknowledgement:} This research was supported in part by NSF Awards IIS-2312865, OAC-2311521 and  IIS-2442137, in addition to a gift to the LinkedIn-Cornell Bowers CIS Strategic Partnership and an AI2050 Early Career Fellowship program at Schmidt Sciences. All content represents the opinion of the
authors, which is not necessarily shared or endorsed by their respective employers and/or sponsors.
\bibliographystyle{abbrvnat}
\bibliography{Bibliography}

\clearpage

\appendix
\section{Appendix}
\label{sec:Appendix}

\subsection{Dataset}
\label{sec:appdata}
Table \ref{tab:datastats} summarizes paper and review counts across venues that we use for analyses. Note that the NeurIPS and ICML data is heavily biased towards accepted papers.
\begin{table}[h]
\centering
\begin{tabular}{lccccccc}
\hline
& \multicolumn{2}{c}{Total} & \multicolumn{2}{c}{Accepted} & \multicolumn{2}{c}{Rejected} \\
\cmidrule(lr){2-3} \cmidrule(lr){4-5} \cmidrule(lr){6-7}
Conference & Papers & Reviews & Papers & Reviews & Papers & Reviews \\
\hline
ICLR 2024 & 7,404 & 28,028 & 2,260 & 8,742 & 5,144 & 19,286 \\
ICLR 2025 & 11,672 & 46,748 & 3,704 & 14,940 & 7,968 & 31,808 \\
NeurIPS 2024 & 4,236 & 16,644 & 4,035 & 15,855 & 201 & 789 \\
NeurIPS 2025 & 5,526 & 22,313 & 5,275 & 21,301 & 251 & 1,012 \\
ICML 2025 & 3,422 & 13,102 & 3,257 & 12,478 & 165 & 624 \\
\hline
\end{tabular}
\caption{Dataset statistics: Submission and review counts across venues.}
\label{tab:datastats}
\end{table}

\subsection{Summary Statistics}
\label{sec:appsummary} Tables~\ref{tab:neurips2024summary},~\ref{tab:neurips2025summary}, and~\ref{tab:icmlfullsummaryapp} present summary statistics for NeurIPS 2024, NeurIPS 2025, and ICML 2025. This includes every field reviewers were required to provide a rating on.

\begin{table}[H]
\centering
\small
\begin{adjustbox}{max width=\textwidth}  
\begin{tabular}{|l|c|c|c|c|c|c|}
\hline
\textbf{Quadrant }\textit{(Paper, Review)} & \textbf{n} & \textbf{Rating} & \textbf{Confidence} & \textbf{Presentation} & \textbf{Contribution} & \textbf{Soundness} \\
\hline
(Human, Human) & 13,020 & $5.87 \pm 1.20$ & $3.60 \pm 0.84$ & $2.86 \pm 0.71$ & $2.72 \pm 0.65$ & $2.92 \pm 0.63$ \\
\hline
(Human, LLM) & 3,432 & $5.86 \pm 1.14$ & $3.62 \pm 0.88$ & $2.91 \pm 0.61$ & $2.78 \pm 0.61$ & $2.94 \pm 0.56$ \\
\hline
(LLM, Human) & 65 & $5.66 \pm 1.33$ & $3.85 \pm 0.73$ & $2.82 \pm 0.61$ & $2.52 \pm 0.66$ & $2.78 \pm 0.62$ \\
\hline
(LLM, LLM) & 58 & $5.97 \pm 1.31$ & $3.78 \pm 0.92$ & $2.84 \pm 0.67$ & $2.79 \pm 0.61$ & $2.95 \pm 0.63$ \\
\hline
\end{tabular}
\end{adjustbox}
\caption{Summary Statistics for (Paper, Review) pairs from NeurIPS 2024. (Paper, Review) pairs are divided into 4 quadrants based on the LLM status of the paper and the review. $\alpha$ value of 0.15 is used as a threshold: Papers and Reviews are classified as LLM-aided if they have $\alpha>0.15$ and Human otherwise. Cell values are presented as Mean $\pm$ Std Dev}
\label{tab:neurips2024summary}
\end{table}

\begin{table}[H]
\centering
\small
\begin{adjustbox}{max width=\textwidth}  
\begin{tabular}{|l|c|c|c|c|c|c|c|}
\hline
\textbf{Quadrant }\textit{(Paper, Review)} & \textbf{n} & \textbf{Rating} & \textbf{Confidence} & \textbf{Quality} & \textbf{Clarity} & \textbf{Significance} & \textbf{Originality} \\
\hline
(Human, Human) & 18,135 & $4.31 \pm 0.74$ & $3.52 \pm 0.82$ & $2.87 \pm 0.61$ & $2.88 \pm 0.70$ & $2.72 \pm 0.65$ & $2.78 \pm 0.64$ \\
\hline
(Human, LLM) & 3,686 & $4.29 \pm 0.68$ & $3.54 \pm 0.84$ & $2.91 \pm 0.56$ & $2.92 \pm 0.61$ & $2.78 \pm 0.63$ & $2.84 \pm 0.62$ \\
\hline
(LLM, Human) & 264 & $4.20 \pm 0.72$ & $3.69 \pm 0.80$ & $2.74 \pm 0.59$ & $2.77 \pm 0.70$ & $2.61 \pm 0.61$ & $2.63 \pm 0.63$ \\
\hline
(LLM, LLM) & 111 & $4.34 \pm 0.69$ & $3.65 \pm 0.87$ & $2.86 \pm 0.60$ & $2.87 \pm 0.63$ & $2.83 \pm 0.57$ & $2.81 \pm 0.60$ \\
\hline
\end{tabular}
\end{adjustbox}
\caption{Summary Statistics for (Paper, Review) pairs from NeurIPS 2025. (Paper, Review) pairs are divided into 4 quadrants based on the LLM status of the paper and the review. $\alpha$ value of 0.15 is used as a threshold: Papers and Reviews are classified as LLM-aided if they have $\alpha>0.15$ and Human otherwise. Cell values are presented as Mean $\pm$ Std Dev}
\label{tab:neurips2025summary}
\end{table}
\begin{table}[H]
\centering
\begin{tabular}{|l|c|}
\hline
\textbf{Quadrant }\textit{(Paper, Review)} & \textbf{Rating} \\
\hline
(Human, Human) & $3.21 \pm 0.80$ \\
\hline
(Human, LLM) & $3.28 \pm 0.76$ \\
\hline
(LLM, Human) & $3.11 \pm 0.84$ \\
\hline
(LLM, LLM) & $3.25 \pm 0.75$ \\
\hline
\end{tabular}
\caption{Summary Statistics for (Paper, Review) pairs from ICML 2025. (Paper, Review) pairs are divided into 4 quadrants based on the LLM status of the paper and the review. $\alpha$ value of 0.15 is used as a threshold: Papers and Reviews are classified as LLM-aided if they have $\alpha>0.15$ and Human otherwise. Cell values are presented as Mean $\pm$ Std Dev. The `Rating' field is on a scale from 1 to 5.}
\label{tab:icmlfullsummaryapp}
\end{table}

\subsection{Detecting LLM Use}
\label{sec:llmdetectionmethod}
We use the method proposed by \citet{liaodetection, liang2024mappingincreasingusellms} to detect the presence of LLM generated text in both papers and reviews. The method frames the detection problem as a Maximum Likelihood Estimation problem as follow:
\begin{enumerate}
    \item Let $P$ and $Q$ denote the probability distribution of documents written by humans and LLMs respectively. Given a new corpus of text $x$, we posit that $x$ was drawn from a weighted sum of these two distributions, i.e., $x$ was generated from the mixture distribution
    \begin{equation*}
        (1-\alpha)P + \alpha Q
    \end{equation*}
    \item The method finds the $\alpha$ that maximizes 
    \begin{equation*}
        \mathcal{L}(\alpha) = \sum_{i=1}^{n} \log((1-\alpha)P(x_i) + \alpha Q(x_i)).
    \end{equation*}
    where $x_1, x_2, \dots x_n$ are sentences in text corpus $x$.
    \item The base distributions $P$ and $Q$ are determined using unigram frequency models estimated from training data. Specifically, $P$ is estimated from human-written text, while $Q$ is estimated from LLM-generated paraphrases of the same text. The authors of \citet{liang2024mappingincreasingusellms} provide pre-computed word frequency distributions trained on computer science paper abstracts from arXiv (categories cs.CL and cs.LG) published before November 2023, paired with their LLM generated paraphrases. We use these provided distributions for our analysis.
\end{enumerate}

\subsection{LLM aided and Human Ratings to papers in different quality bins}
\begin{figure}[h]
\centering
\includegraphics[width=0.8\textwidth]{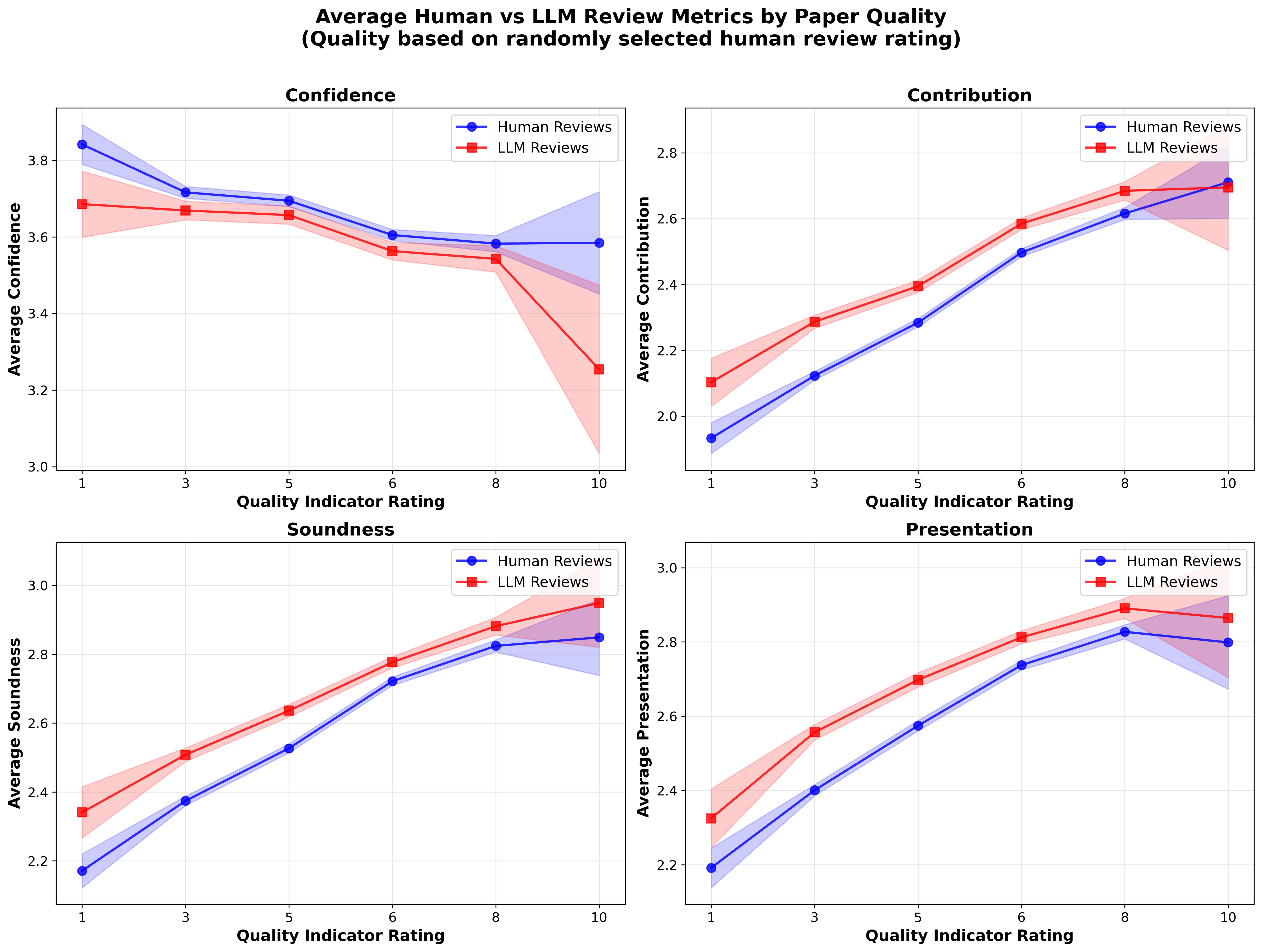}
\caption{Average Rating provided by Human reviews and LLM-aided reviews to Papers bucketed by paper quality. Clockwise from top-left, we have Confidence, Contribution, Soundness, and Presentation.}
\label{fig:llmvshumanbinapp}
\end{figure}

Figure \ref{fig:llmvshumanbinapp} shows the difference in average scores given by LLM-aided reviews and Human reviews to papers in different quality bins. We only consider papers that received at least one human review. Here, quality is determined by using a left out human review received by the paper, and the remaining human reviews are then used to plot the human curve on the graph. \textit{Contribution, Soundness and Presentation} follow the same trend we saw Rating follow in Figure \ref{fig:llm_leniency_by_quality}: LLM-aided reviews give slightly higher scores to low quality papers than human reviews but they give similar scores to medium and high quality papers. The \textit{Confidence} trend lines are interesting: both LLM-aided reviews and human reviews show lower confidence when providing ratings to higher quality papers. The confidence scores given by human reviews are slightly higher than those given by LLM-aided reviews, one speculative reason for which could be these reviewers manually assigning a lower confidence rating since they used an LLM to help with the review.
\subsection{Fully LLM and Human Ratings to papers in different quality bins}
\begin{figure}[h]
\centering
\includegraphics[width=0.8\textwidth]{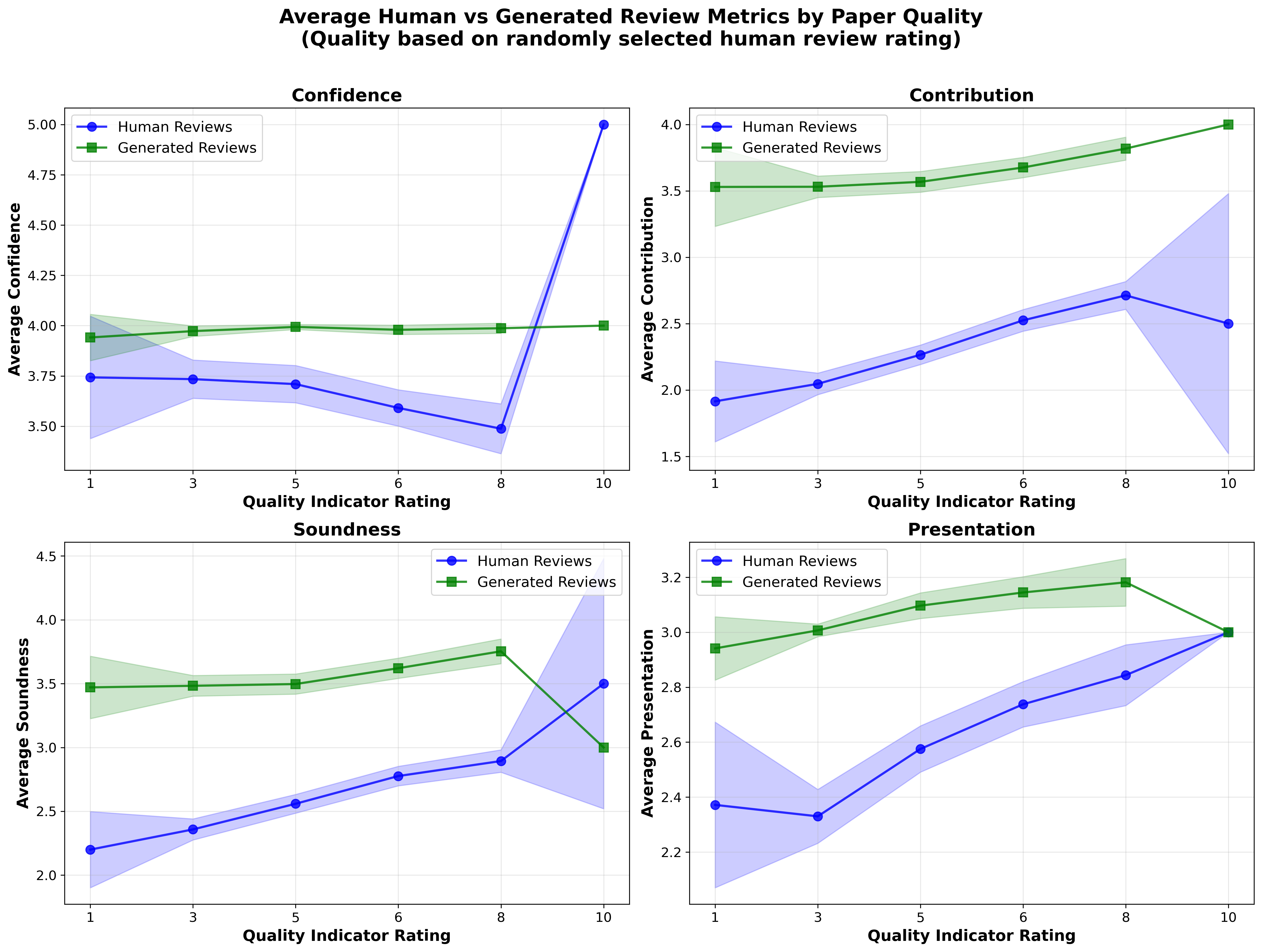}
\caption{Average Rating provided by Human reviews and LLM-aided reviews to Papers ordered by their quality percentile. Clockwise from top-left, we have Confidence, Contribution, Soundness, and Presentation.}
\label{fig:fullyllmvshumanbinapp}
\end{figure}

Figure \ref{fig:fullyllmvshumanbinapp} shows the difference in average scores given by fully LLM generated reviews and Human reviews to papers in different quality bins. Here, quality is determined by using a left out human review received by the paper, and the remaining human reviews are then used to plot the human curve on the graph. Again, \textit{Contribution, Soundness and Presentation} follow the same trend we saw Rating follow in Figure \ref{fig:generatelines}. The fully LLM generated reviews give substantially higher scores to papers in every quality bin and the generated review curve in these three graphs barely has a positive slope. Meanwhile, these reviews seem to uniformly assign a confidence score of 4 to every rating, again implying the reviewers using an LLM post-hoc lowering the confidence score they provide in their review. Note that the quality bucket with rating `10' only has a one paper, which is why the curve trend abruptly deviates at that point.

\subsection{Full Regression Analyses}
\subsubsection{ICLR}

\textit{Regression with area correction}: 
Using the framing from Equation \ref{eq:regression}
\begin{equation}
Y_{ij} = \beta_0 + \beta_1 X_i^{\text{LLM}} + \beta_2 T_{ij} + \beta_3 (X_i^{\text{LLM}} \times T_{ij}) + \gamma' Z_i + \epsilon_{ij}
\end{equation}
where:
\begin{itemize}
    \item $X_i^{\text{LLM}} \in \{0, 1\}$ is a binary indicator that indicates whether paper $i$ was written with LLM assistance
    \item $T_{ij} \in \{0, 1\}$ is a binary indicator that indicates whether review $j$ was written with LLM assistance (the treatment)
    \item $Z_i$ is a vector of control variables (e.g., area fixed effects). These are the confounders we control for by including them in our regression analysis. For our regression, this is a group of indicator variables that take a value of 1 if the paper belongs to a particular area and 0 otherwise.
    \item $\epsilon_{ij}$ is the noise term
\end{itemize}

For ICLR 2024, there are 20 unique research areas and we add an indicator term for all but one reference area in our regression formulation. For ICLR 2025, there are 21 unique research areas. The overlap between paper areas across the two years is not clean, requiring us to develop a mapping to combine the datasets. After mapping equivalent areas, we are left with 23 unique research areas across both years.

Table~\ref{tab:iclr_areas} shows the research areas for each year and the combined dataset. The four area mappings we apply are: (1) ``general machine learning'' (2024) maps to ``other topics in machine learning'' (2025), (2) ``societal considerations including fairness/safety/privacy'' (2024) maps to ``alignment/fairness/safety/privacy/societal considerations'' (2025), (3) ``infrastructure, software libraries, hardware, etc.'' (2024) maps to ``infrastructure, software libraries, hardware, systems, etc.'' (2025), and (4) ``representation learning for computer vision/audio/language/other modalities'' (2024) maps to ``applications to computer vision/audio/language/other modalities'' (2025).

\begin{table}[h]
\centering
\small
\begin{tabular}{|p{0.3\textwidth}|p{0.3\textwidth}|p{0.3\textwidth}|}
\hline
\textbf{ICLR 2024 (20 areas)} & \textbf{ICLR 2025 (21 areas)} & \textbf{Combined (23 areas)} \\
\hline
Causal reasoning & Causal reasoning & Causal reasoning \\
Datasets \& benchmarks & Datasets \& benchmarks & Datasets \& benchmarks \\
General ML & Other topics in ML & General/other ML \\
Generative models & Generative models & Generative models \\
Graph \& geometry learning & Graph \& geometry learning & Graph \& geometry learning \\
Infrastructure & Infrastructure & Infrastructure \\
--- & Foundation models & Foundation models \\
--- & Interpretability \& explainable AI & Interpretability \& explainable AI \\
Learning theory & Learning theory & Learning theory \\
Metric/kernel/sparse coding & --- & Metric/kernel/sparse coding \\
Representation learning (CV/audio/language) & Multimodal applications & Multimodal applications \\
Neuroscience \& cognitive & Neuroscience \& cognitive & Neuroscience \& cognitive \\
Neurosymbolic \& hybrid & Neurosymbolic \& hybrid & Neurosymbolic \& hybrid \\
Optimization & Optimization & Optimization \\
Physical sciences & Physical sciences & Physical sciences \\
Probabilistic methods & Probabilistic methods & Probabilistic methods \\
Reinforcement learning & Reinforcement learning & Reinforcement learning \\
Representation learning & Representation learning & Representation learning \\
Robotics \& autonomy & Robotics \& autonomy & Robotics \& autonomy \\
Societal considerations & Alignment/fairness/safety/privacy & Societal considerations \\
--- & Time series \& dynamical systems & Time series \& dynamical systems \\
Transfer/meta/lifelong & Transfer/meta/lifelong & Transfer/meta/lifelong \\
Visualization/interpretation & --- & Visualization/interpretation \\
\hline
\end{tabular}
\caption{Research areas for ICLR 2024, ICLR 2025, and the combined dataset. Areas unique to one year are marked with ``---'' in the other column. Four area pairs were mapped together for the combined analysis (see text).}
\label{tab:iclr_areas}
\end{table}

\paragraph{ICLR 2024 Results}
For ICLR 2024, we analyze 27,859 (paper, review) pairs. Among these, 743 papers (2.7\%) are classified as LLM-aided, 4,327 reviews (15.5\%) are classified as LLM-aided, and 168 pairs (0.6\%) have both paper and review classified as LLM-aided.

Table~\ref{tab:iclr2024_main_regression} presents the regression coefficients for the main variables. The intercept $\beta_0 = 5.114$ ($p < 0.0001$) represents the baseline rating for human-written papers with human-written reviews in the baseline area (causal reasoning). The coefficient $\beta_1 = -1.058$ ($p < 0.0001$) indicates that LLM-aided papers receive substantially lower ratings on average, holding review type and area constant.

The CATE estimates reveal an asymmetric relationship. For human-written papers (CATE(Human) = $\beta_2$), LLM-aided reviews increase ratings by $0.271$ points ($p < 0.0001$, 95\% CI: [0.215, 0.328]). For LLM-aided papers (CATE(LLM) = $\beta_2 + \beta_3$), LLM-aided reviews increase ratings by $0.615$ points ($p < 0.0001$, 95\% CI: [0.322, 0.909]). The difference in treatment effects ($\beta_3 = 0.344$, $p = 0.024$, 95\% CI: [0.046, 0.643]) indicates that LLM-aided reviews have a significantly stronger positive effect when evaluating LLM-aided papers.

\begin{table}[h]
\centering
\begin{tabular}{|l|c|c|c|c|}
\hline
\textbf{Parameter} & \textbf{Coefficient} & \textbf{Std Error} & \textbf{t-statistic} & \textbf{p-value} \\
\hline
$\beta_0$ (Intercept) & 5.114 & 0.086 & 59.32 & $<0.0001$ \\
$\beta_1$ (Paper LLM) & $-1.058$ & 0.072 & $-14.66$ & $<0.0001$ \\
$\beta_2$ (Review LLM) & 0.271 & 0.029 & 9.40 & $<0.0001$ \\
$\beta_3$ (Both LLM) & 0.344 & 0.152 & 2.26 & 0.024 \\
\hline
\multicolumn{5}{|c|}{\textit{CATE Estimates}} \\
\hline
CATE(Human) = $\beta_2$ & 0.271 & 0.029 & 9.40 & $<0.0001$ \\
CATE(LLM) = $\beta_2 + \beta_3$ & 0.615 & 0.150 & 4.11 & $<0.0001$ \\
Difference = $\beta_3$ & 0.344 & 0.152 & 2.26 & 0.024 \\
\hline
\end{tabular}
\caption{Main regression results for ICLR 2024 (N = 27,859). CATE(Human) represents the effect of LLM-aided reviews on human-written papers. CATE(LLM) represents the effect of LLM-aided reviews on LLM-aided papers.}
\label{tab:iclr2024_main_regression}
\end{table}

"Signficant" area coefficients are presented in Table~\ref{tab:iclr2024_area_coefficients}. Among the area effects, learning theory shows the strongest positive effect ($0.583$, $p < 0.0001$), while transfer/meta/lifelong learning shows a significant negative effect ($-0.232$, $p = 0.017$). Generative models ($0.203$, $p = 0.027$) and probabilistic methods ($0.223$, $p = 0.037$) also show significant positive effects relative to the baseline.

\begin{table}[h]
\centering
\begin{tabular}{|l|c|c|c|c|}
\hline
\textbf{Area} & \textbf{Coefficient} & \textbf{Std Error} & \textbf{t-statistic} & \textbf{p-value} \\
\hline
Causal Reasoning (baseline) & 0.000 & --- & --- & --- \\
Generative Models & 0.203 & 0.092 & 2.22 & 0.027 \\
Learning Theory & 0.583 & 0.105 & 5.54 & $<0.0001$ \\
Probabilistic Methods & 0.223 & 0.107 & 2.09 & 0.037 \\
Transfer/Meta/Lifelong & $-0.232$ & 0.097 & $-2.39$ & 0.017 \\
\hline
\end{tabular}
\caption{Significant area coefficients for ICLR 2024 regression (N = 27,859). Only areas with $p < 0.05$ are shown. All coefficients are relative to the baseline (causal reasoning).}
\label{tab:iclr2024_area_coefficients}
\end{table}

\paragraph{ICLR 2025 Results}
For ICLR 2025, we analyze 46,555 (paper, review) pairs. Among these, 1,482 papers (3.2\%) are classified as LLM-aided, 12,417 reviews (26.7\%) are classified as LLM-aided, and 517 pairs (1.1\%) have both paper and review classified as LLM-aided.

Table~\ref{tab:iclr2025_main_regression} presents the regression results. The intercept $\beta_0 = 5.168$ ($p < 0.0001$) is slightly higher than 2024. The coefficient for LLM-aided papers is $\beta_1 = -0.902$ ($p < 0.0001$), somewhat smaller in magnitude than 2024 but still highly significant and negative.

The CATE estimates show similar patterns to 2024. For human-written papers, LLM-aided reviews increase ratings by $0.257$ points ($p < 0.0001$, 95\% CI: [0.221, 0.292]). For LLM-aided papers, LLM-aided reviews increase ratings by $0.617$ points ($p < 0.0001$, 95\% CI: [0.436, 0.798]). The difference in treatment effects is $\beta_3 = 0.360$ ($p = 0.0001$, 95\% CI: [0.175, 0.545]), confirming the asymmetric effect found in 2024.

\begin{table}[h]
\centering
\begin{tabular}{|l|c|c|c|c|}
\hline
\textbf{Parameter} & \textbf{Coefficient} & \textbf{Std Error} & \textbf{t-statistic} & \textbf{p-value} \\
\hline
$\beta_0$ (Intercept) & 5.168 & 0.078 & 66.11 & $<0.0001$ \\
$\beta_1$ (Paper LLM) & $-0.902$ & 0.055 & $-16.27$ & $<0.0001$ \\
$\beta_2$ (Review LLM) & 0.257 & 0.018 & 14.09 & $<0.0001$ \\
$\beta_3$ (Both LLM) & 0.360 & 0.094 & 3.82 & 0.0001 \\
\hline
\multicolumn{5}{|c|}{\textit{CATE Estimates}} \\
\hline
CATE(Human) = $\beta_2$ & 0.257 & 0.018 & 14.09 & $<0.0001$ \\
CATE(LLM) = $\beta_2 + \beta_3$ & 0.617 & 0.092 & 6.68 & $<0.0001$ \\
Difference = $\beta_3$ & 0.360 & 0.094 & 3.82 & 0.0001 \\
\hline
\end{tabular}
\caption{Main regression results for ICLR 2025 (N = 46,555). CATE(Human) represents the effect of LLM-aided reviews on human-written papers. CATE(LLM) represents the effect of LLM-aided reviews on LLM-aided papers.}
\label{tab:iclr2025_main_regression}
\end{table}

"Significant" area coefficients are presented in Table~\ref{tab:iclr2025_area_coefficients}. Among area coefficients, learning theory again shows the strongest positive effect ($0.266$, $p = 0.003$). Transfer/meta/lifelong learning ($-0.370$, $p < 0.0001$), time series \& dynamical systems ($-0.279$, $p = 0.003$), representation learning ($-0.246$, $p = 0.003$), and general ML ($-0.243$, $p = 0.005$) all show significant negative effects relative to the baseline.

\begin{table}[h]
\centering
\begin{tabular}{|l|c|c|c|c|}
\hline
\textbf{Area} & \textbf{Coefficient} & \textbf{Std Error} & \textbf{t-statistic} & \textbf{p-value} \\
\hline
Causal Reasoning (baseline) & 0.000 & --- & --- & --- \\
General ML & $-0.243$ & 0.086 & $-2.83$ & 0.005 \\
Learning Theory & 0.266 & 0.090 & 2.96 & 0.003 \\
Representation Learning & $-0.246$ & 0.083 & $-2.95$ & 0.003 \\
Time Series \& Dynamical & $-0.279$ & 0.094 & $-2.98$ & 0.003 \\
Transfer/Meta/Lifelong & $-0.370$ & 0.090 & $-4.13$ & $<0.0001$ \\
\hline
\end{tabular}
\caption{Significant area coefficients for ICLR 2025 regression (N = 46,555). Only areas with $p < 0.05$ are shown. All coefficients are relative to the baseline (causal reasoning).}
\label{tab:iclr2025_area_coefficients}
\end{table}

\paragraph{Combined ICLR 2024-2025 Results}
Combining both years yields 74,414 (paper, review) pairs. Among these, 2,225 papers (3.0\%) are classified as LLM-aided, 16,744 reviews (22.5\%) are classified as LLM-aided, and 685 pairs (0.9\%) have both paper and review classified as LLM-aided.

Table~\ref{tab:iclr_combined_main_regression} presents the regression results for the combined dataset. The coefficient for LLM-aided papers is $\beta_1 = -0.962$ ($p < 0.0001$), falling between the 2024 and 2025 values. The CATE estimates show LLM-aided reviews increase ratings by $0.253$ points for human-written papers ($p < 0.0001$, 95\% CI: [0.223, 0.283]) and $0.629$ points for LLM-aided papers ($p < 0.0001$, 95\% CI: [0.476, 0.782]). The difference in treatment effects is $\beta_3 = 0.376$ ($p < 0.0001$, 95\% CI: [0.221, 0.532]).

\begin{table}[h]
\centering
\begin{tabular}{|l|c|c|c|c|}
\hline
\textbf{Parameter} & \textbf{Coefficient} & \textbf{Std Error} & \textbf{t-statistic} & \textbf{p-value} \\
\hline
$\beta_0$ (Intercept) & 5.145 & 0.058 & 88.87 & $<0.0001$ \\
$\beta_1$ (Paper LLM) & $-0.962$ & 0.044 & $-21.87$ & $<0.0001$ \\
$\beta_2$ (Review LLM) & 0.253 & 0.015 & 16.52 & $<0.0001$ \\
$\beta_3$ (Both LLM) & 0.376 & 0.080 & 4.74 & $<0.0001$ \\
\hline
\multicolumn{5}{|c|}{\textit{CATE Estimates}} \\
\hline
CATE(Human) = $\beta_2$ & 0.253 & 0.015 & 16.52 & $<0.0001$ \\
CATE(LLM) = $\beta_2 + \beta_3$ & 0.629 & 0.078 & 8.06 & $<0.0001$ \\
Difference = $\beta_3$ & 0.376 & 0.080 & 4.74 & $<0.0001$ \\
\hline
\end{tabular}
\caption{Main regression results for Combined ICLR 2024-2025 (N = 74,414). CATE(Human) represents the effect of LLM-aided reviews on human-written papers. CATE(LLM) represents the effect of LLM-aided reviews on LLM-aided papers.}
\label{tab:iclr_combined_main_regression}
\end{table}

Area coefficients are presented in Table~\ref{tab:iclr_combined_area_coefficients}. The combined analysis shows learning theory with the strongest positive effect ($0.383$, $p < 0.0001$), while transfer/meta/lifelong learning ($-0.304$, $p < 0.0001$), time series \& dynamical systems ($-0.251$, $p = 0.001$), representation learning ($-0.200$, $p = 0.001$), and general ML ($-0.163$, $p = 0.009$) show significant negative effects. Generative models shows a modest positive effect ($0.128$, $p = 0.037$).

\begin{table}[h]
\centering
\begin{tabular}{|l|c|c|c|c|}
\hline
\textbf{Area} & \textbf{Coefficient} & \textbf{Std Error} & \textbf{t-statistic} & \textbf{p-value} \\
\hline
Causal Reasoning (baseline) & 0.000 & --- & --- & --- \\
General ML & $-0.163$ & 0.062 & $-2.61$ & 0.009 \\
Generative Models & 0.128 & 0.061 & 2.09 & 0.037 \\
Learning Theory & 0.383 & 0.068 & 5.62 & $<0.0001$ \\
Representation Learning & $-0.200$ & 0.061 & $-3.26$ & 0.001 \\
Time Series \& Dynamical & $-0.251$ & 0.078 & $-3.24$ & 0.001 \\
Transfer/Meta/Lifelong & $-0.304$ & 0.066 & $-4.62$ & $<0.0001$ \\
\hline
\end{tabular}
\caption{Significant area coefficients for Combined ICLR 2024-2025 regression (N = 74,414). Only areas with $p < 0.05$ are shown. All coefficients are relative to the baseline (causal reasoning).}
\label{tab:iclr_combined_area_coefficients}
\end{table}

\paragraph{Accepted Papers Only}
Meanwhile, when we only consider accepted papers, the patterns change substantially. For ICLR 2024 accepted papers, we analyze 8,706 (paper, review) pairs. Among these, only 96 papers (1.1\%) are classified as LLM-aided, 1,447 reviews (16.6\%) are classified as LLM-aided, and 31 pairs (0.4\%) have both paper and review classified as LLM-aided. For ICLR 2025 accepted papers, we analyze 14,884 pairs, with 349 papers (2.3\%) classified as LLM-aided, 4,092 reviews (27.5\%) classified as LLM-aided, and 133 pairs (0.9\%) with both classified as LLM-aided. The combined accepted papers dataset contains 23,590 pairs, with 445 papers (1.9\%) classified as LLM-aided, 5,539 reviews (23.5\%) classified as LLM-aided, and 164 pairs (0.7\%) with both classified as LLM-aided.

Table~\ref{tab:iclr_accepted_main_regression} presents the regression results for accepted papers across all three datasets. Notably, the coefficient for LLM-aided papers ($\beta_1$) is no longer significantly negative. For ICLR 2024 accepted papers, $\beta_1 = 0.217$ ($p = 0.188$); for ICLR 2025, $\beta_1 = -0.045$ ($p = 0.609$); and for the combined dataset, $\beta_1 = 0.012$ ($p = 0.880$). None of these coefficients are statistically significant, indicating that among accepted papers, LLM-aided papers do not receive systematically different ratings than human-written papers after controlling for research area.

The CATE estimates for accepted papers differ markedly from the full dataset. For ICLR 2024 accepted papers, neither CATE(Human) ($-0.016$, $p = 0.676$, 95\% CI: [-0.091, 0.059]) nor CATE(LLM) ($-0.102$, $p = 0.724$, 95\% CI: [-0.665, 0.462]) is statistically significant. For ICLR 2025 accepted papers, CATE(Human) is small but significant ($-0.063$, $p = 0.009$, 95\% CI: [-0.110, -0.016]), while CATE(LLM) is not significant ($0.140$, $p = 0.324$, 95\% CI: [-0.138, 0.418]). The difference in treatment effects ($\beta_3$) is not significant in either year: $-0.086$ for 2024 ($p = 0.768$) and $0.203$ for 2025 ($p = 0.158$).

For the combined accepted papers dataset, CATE(Human) is small but significant ($-0.051$, $p = 0.011$, 95\% CI: [-0.091, -0.012]), indicating that LLM-aided reviews are associated with slightly lower ratings for human-written papers among those that were accepted. However, CATE(LLM) is not significant ($0.082$, $p = 0.520$, 95\% CI: [-0.168, 0.332]), and the difference in treatment effects ($\beta_3 = 0.134$, $p = 0.301$, 95\% CI: [-0.120, 0.387]) is also not significant.

\begin{table}[h]
\centering
\small
\begin{tabular}{|l|c|c|c|c|c|}
\hline
\textbf{Dataset} & \textbf{N} & \textbf{$\beta_0$} & \textbf{$\beta_1$} & \textbf{$\beta_2$} & \textbf{$\beta_3$} \\
\hline
\multicolumn{6}{|c|}{\textit{Main Coefficients}} \\
\hline
ICLR 2024 (Accepted) & 8,706 & $6.528***$ & $0.217$ & $-0.016$ & $-0.086$ \\
 & & $(0.120)$ & $(0.165)$ & $(0.038)$ & $(0.290)$ \\
\hline
ICLR 2025 (Accepted) & 14,884 & $6.556***$ & $-0.045$ & $-0.063**$ & $0.203$ \\
 & & $(0.106)$ & $(0.089)$ & $(0.024)$ & $(0.144)$ \\
\hline
Combined (Accepted) & 23,590 & $6.544***$ & $0.012$ & $-0.051*$ & $0.134$ \\
 & & $(0.079)$ & $(0.078)$ & $(0.020)$ & $(0.129)$ \\
\hline
\multicolumn{6}{|c|}{\textit{CATE Estimates}} \\
\hline
\multicolumn{6}{|l|}{ICLR 2024 (Accepted):} \\
\quad CATE(Human) = $\beta_2$ & \multicolumn{5}{l|}{$-0.016$ (SE: $0.038$, $p = 0.676$, 95\% CI: [-0.091, 0.059])} \\
\quad CATE(LLM) = $\beta_2 + \beta_3$ & \multicolumn{5}{l|}{$-0.102$ (SE: $0.287$, $p = 0.724$, 95\% CI: [-0.665, 0.462])} \\
\hline
\multicolumn{6}{|l|}{ICLR 2025 (Accepted):} \\
\quad CATE(Human) = $\beta_2$ & \multicolumn{5}{l|}{$-0.063$ (SE: $0.024$, $p = 0.009$, 95\% CI: [-0.110, -0.016])} \\
\quad CATE(LLM) = $\beta_2 + \beta_3$ & \multicolumn{5}{l|}{$0.140$ (SE: $0.142$, $p = 0.324$, 95\% CI: [-0.138, 0.418])} \\
\hline
\multicolumn{6}{|l|}{Combined (Accepted):} \\
\quad CATE(Human) = $\beta_2$ & \multicolumn{5}{l|}{$-0.051$ (SE: $0.020$, $p = 0.011$, 95\% CI: [-0.091, -0.012])} \\
\quad CATE(LLM) = $\beta_2 + \beta_3$ & \multicolumn{5}{l|}{$0.082$ (SE: $0.128$, $p = 0.520$, 95\% CI: [-0.168, 0.332])} \\
\hline
\end{tabular}
\caption{Regression results for ICLR accepted papers only. Standard errors in parentheses. $***p<0.001$, $**p<0.01$, $*p<0.05$. All regressions control for research area fixed effects. The reduction in statistical significance compared to the full dataset (Table~\ref{tab:iclr_combined_main_regression}) suggests that the patterns observed in the full data are driven primarily by rejected papers.}
\label{tab:iclr_accepted_main_regression}
\end{table}

These results indicate that the strong negative effect of LLM-aided papers and the asymmetric positive effect of LLM-aided reviews observed in the full dataset are largely driven by rejected papers. Among accepted papers, LLM usage in papers does not predict significantly different ratings, and the effect of LLM-aided reviews is either negligible or slightly negative. The asymmetric treatment effect that was prominent in the full dataset disappears when restricting to accepted papers.

We also find that most area effects are not statistically significant in the accepted papers analysis. For ICLR 2024 accepted papers, none of the area coefficients reach statistical significance. For ICLR 2025 accepted papers, only transfer/meta/lifelong learning shows a significant negative effect ($-0.303$, $p = 0.015$). For the combined accepted papers dataset, transfer/meta/lifelong learning shows a marginally significant negative effect ($-0.173$, $p = 0.063$). We omit these tables for brevity.

\subsubsection{NeurIPS}
\label{sec:appregneu}
For NeurIPS 2024, there are 31 unique research areas and we add an indicator term for all but one reference area in our regression formulation. For NeurIPS 2025, there are 13 unique research areas. The area taxonomies differ substantially between years, preventing direct combination of the datasets. We therefore analyze each year separately and do not create a combined NeurIPS model.

\paragraph{NeurIPS 2024 Results}
For NeurIPS 2024, we analyze 16,575 (paper, review) pairs. Among these, 123 papers (0.7\%) are classified as LLM-aided, 3,467 reviews (20.9\%) are classified as LLM-aided, and 58 pairs (0.3\%) have both paper and review classified as LLM-aided.

Table~\ref{tab:neurips2024_main_regressionapp} presents the regression results. The intercept $\beta_0 = 5.782$ ($p < 0.0001$) represents the baseline rating for human-written papers with human-written reviews in the baseline area (active learning). The coefficient for LLM-aided papers is $\beta_1 = -0.148$ ($p = 0.316$), which is negative but not statistically significant.

The CATE estimates show no significant effects. For human-written papers, LLM-aided reviews increase ratings by $0.022$ points ($p = 0.340$, 95\% CI: [-0.023, 0.067]). For LLM-aided papers, LLM-aided reviews increase ratings by $0.303$ points ($p = 0.156$, 95\% CI: [-0.115, 0.720]). Neither effect is statistically significant. The difference in treatment effects is $\beta_3 = 0.281$ ($p = 0.190$, 95\% CI: [-0.139, 0.701]), also not statistically significant.

\begin{table}[h]
\centering
\begin{tabular}{|l|c|c|c|c|}
\hline
\textbf{Parameter} & \textbf{Coefficient} & \textbf{Std Error} & \textbf{t-statistic} & \textbf{p-value} \\
\hline
$\beta_0$ (Intercept) & 5.782 & 0.121 & 47.95 & $<0.0001$ \\
$\beta_1$ (Paper LLM) & $-0.148$ & 0.147 & $-1.00$ & 0.316 \\
$\beta_2$ (Review LLM) & 0.022 & 0.023 & 0.95 & 0.340 \\
$\beta_3$ (Interaction) & 0.281 & 0.214 & 1.31 & 0.190 \\
\hline
\multicolumn{5}{|c|}{\textit{CATE Estimates}} \\
\hline
CATE(Human) = $\beta_2$ & 0.022 & 0.023 & 0.95 & 0.340 \\
CATE(LLM) = $\beta_2 + \beta_3$ & 0.303 & 0.213 & 1.42 & 0.156 \\
Difference = $\beta_3$ & 0.281 & 0.214 & 1.31 & 0.190 \\
\hline
\end{tabular}
\caption{Main regression results for NeurIPS 2024 (N = 16,575). CATE(Human) represents the effect of LLM-aided reviews on human-written papers. CATE(LLM) represents the effect of LLM-aided reviews on LLM-aided papers.}
\label{tab:neurips2024_main_regressionapp}
\end{table}

Significant area coefficients are presented in Table~\ref{tab:neurips2024_area_coefficientsapp}. Among the area effects, neuroscience and cognitive science shows the strongest positive effect ($0.313$, $p = 0.019$), followed by learning theory ($0.306$, $p = 0.015$). Algorithmic game theory ($0.301$, $p = 0.037$) and probabilistic methods ($0.282$, $p = 0.031$) also show significant positive effects relative to the baseline.

\begin{table}[h]
\centering
\begin{tabular}{|l|c|c|c|c|}
\hline
\textbf{Area} & \textbf{Coefficient} & \textbf{Std Error} & \textbf{t-statistic} & \textbf{p-value} \\
\hline
Active Learning (baseline) & 0.000 & --- & --- & --- \\
Algorithmic Game Theory & 0.301 & 0.144 & 2.09 & 0.037 \\
Learning Theory & 0.306 & 0.126 & 2.43 & 0.015 \\
Neuroscience \& Cognitive Science & 0.313 & 0.133 & 2.35 & 0.019 \\
Probabilistic Methods & 0.282 & 0.131 & 2.15 & 0.031 \\
\hline
\end{tabular}
\caption{Significant area coefficients for NeurIPS 2024 regression (N = 16,575). Only areas with $p < 0.05$ are shown. All coefficients are relative to the baseline (active learning).}
\label{tab:neurips2024_area_coefficientsapp}
\end{table}

\paragraph{NeurIPS 2025 Results}
For NeurIPS 2025, we analyze 22,196 (paper, review) pairs. Among these, 367 papers (1.7\%) are classified as LLM-aided, 3,750 reviews (16.9\%) are classified as LLM-aided, and 107 pairs (0.5\%) have both paper and review classified as LLM-aided.

Table~\ref{tab:neurips2025_main_regressionapp} presents the regression results. The intercept $\beta_0 = 4.291$ ($p < 0.0001$) represents the baseline rating for human-written papers with human-written reviews in the baseline area (applications). The coefficient for LLM-aided papers is $\beta_1 = -0.081$ ($p = 0.074$), which is negative and marginally significant.

The CATE estimates reveal an interesting pattern. For human-written papers, LLM-aided reviews have essentially no effect ($-0.006$ points, $p = 0.663$, 95\% CI: [-0.032, 0.020]). For LLM-aided papers, LLM-aided reviews increase ratings by $0.177$ points ($p = 0.034$, 95\% CI: [0.013, 0.341]), a statistically significant effect. The difference in treatment effects is $\beta_3 = 0.183$ ($p = 0.031$, 95\% CI: [0.017, 0.348]), indicating that LLM-aided reviews have a significantly stronger positive effect when evaluating LLM-aided papers compared to human-written papers.

\begin{table}[h]
\centering
\begin{tabular}{|l|c|c|c|c|}
\hline
\textbf{Parameter} & \textbf{Coefficient} & \textbf{Std Error} & \textbf{t-statistic} & \textbf{p-value} \\
\hline
$\beta_0$ (Intercept) & 4.291 & 0.012 & 362.95 & $<0.0001$ \\
$\beta_1$ (Paper LLM) & $-0.081$ & 0.046 & $-1.79$ & 0.074 \\
$\beta_2$ (Review LLM) & $-0.006$ & 0.013 & $-0.44$ & 0.663 \\
$\beta_3$ (Interaction) & 0.183 & 0.085 & 2.16 & 0.031 \\
\hline
\multicolumn{5}{|c|}{\textit{CATE Estimates}} \\
\hline
CATE(Human) = $\beta_2$ & $-0.006$ & 0.013 & $-0.44$ & 0.663 \\
CATE(LLM) = $\beta_2 + \beta_3$ & 0.177 & 0.083 & 2.12 & 0.034 \\
Difference = $\beta_3$ & 0.183 & 0.085 & 2.16 & 0.031 \\
\hline
\end{tabular}
\caption{Main regression results for NeurIPS 2025 (N = 22,196). CATE(Human) represents the effect of LLM-aided reviews on human-written papers. CATE(LLM) represents the effect of LLM-aided reviews on LLM-aided papers.}
\label{tab:neurips2025_main_regressionapp}
\end{table}

Significant area coefficients are presented in Table~\ref{tab:neurips2025_area_coefficientsapp}. Among area coefficients, theory shows the strongest positive effect ($0.128$, $p < 0.0001$), followed by probabilistic methods ($0.070$, $p = 0.010$) and neuroscience and cognitive science ($0.069$, $p = 0.023$). The area labeled "other" shows a significant negative effect ($-0.143$, $p = 0.003$).

\begin{table}[h]
\centering
\begin{tabular}{|l|c|c|c|c|}
\hline
\textbf{Area} & \textbf{Coefficient} & \textbf{Std Error} & \textbf{t-statistic} & \textbf{p-value} \\
\hline
Applications (baseline) & 0.000 & --- & --- & --- \\
Neuroscience \& Cognitive Science & 0.069 & 0.030 & 2.28 & 0.023 \\
Other & $-0.143$ & 0.048 & $-3.00$ & 0.003 \\
Probabilistic Methods & 0.070 & 0.027 & 2.58 & 0.010 \\
Theory & 0.128 & 0.020 & 6.44 & $<0.0001$ \\
\hline
\end{tabular}
\caption{Significant area coefficients for NeurIPS 2025 regression (N = 22,196). Only areas with $p < 0.05$ are shown. All coefficients are relative to the baseline (applications).}
\label{tab:neurips2025_area_coefficientsapp}
\end{table}

\subsubsection{ICML}
\label{sec:appregicml}
For ICML 2025, there are 8 unique research areas and we again add an indicator term for all but one reference area in our regression formulation. The research areas are: applications, deep learning, general machine learning, optimization, probabilistic methods, reinforcement learning, social aspects, and theory.

We analyze 13,063 (paper, review) pairs from ICML 2025. Among these, 292 papers (2.2\%) are classified as LLM-aided, 2,451 reviews (18.8\%) are classified as LLM-aided, and 110 pairs (0.8\%) have both paper and review classified as LLM-aided.

Table~\ref{tab:icml2025_main_regression} presents the regression results. The intercept $\beta_0 = 3.181$ ($p < 0.0001$) represents the baseline rating for human-written papers with human-written reviews in the baseline area (applications). The coefficient for LLM-aided papers is $\beta_1 = -0.095$ ($p = 0.111$), which is negative but not statistically significant, indicating that LLM-aided papers do not receive systematically different ratings after controlling for research area and review type.

For human-written papers, LLM-aided reviews increase ratings by $0.079$ points ($p < 0.0001$, 95\% CI: [0.043, 0.115]). For LLM-aided papers, LLM-aided reviews increase ratings by $0.164$ points ($p = 0.089$, 95\% CI: [-0.025, 0.354]), though this effect is only marginally significant. The difference in treatment effects is $\beta_3 = 0.085$ ($p = 0.386$, 95\% CI: [-0.107, 0.278]), which is not statistically significant, suggesting that unlike ICLR, the effect of LLM-aided reviews does not differ significantly between human-written and LLM-aided papers in ICML.

\begin{table}[h]
\centering
\begin{tabular}{|l|c|c|c|c|}
\hline
\textbf{Parameter} & \textbf{Coefficient} & \textbf{Std Error} & \textbf{t-statistic} & \textbf{p-value} \\
\hline
$\beta_0$ (Intercept) & 3.181 & 0.017 & 188.21 & $<0.0001$ \\
$\beta_1$ (Paper LLM) & $-0.095$ & 0.060 & $-1.59$ & 0.111 \\
$\beta_2$ (Review LLM) & 0.079 & 0.018 & 4.33 & $<0.0001$ \\
$\beta_3$ (Interaction) & 0.085 & 0.098 & 0.87 & 0.386 \\
\hline
\multicolumn{5}{|c|}{\textit{CATE Estimates}} \\
\hline
CATE(Human) = $\beta_2$ & 0.079 & 0.018 & 4.33 & $<0.0001$ \\
CATE(LLM) = $\beta_2 + \beta_3$ & 0.164 & 0.097 & 1.70 & 0.089 \\
Difference = $\beta_3$ & 0.085 & 0.098 & 0.87 & 0.386 \\
\hline
\end{tabular}
\caption{Main regression results for ICML 2025 (N = 13,063). CATE(Human) represents the effect of LLM-aided reviews on human-written papers. CATE(LLM) represents the effect of LLM-aided reviews on LLM-aided papers.}
\label{tab:icml2025_main_regression}
\end{table}

Area coefficients are presented in Table~\ref{tab:icml2025_area_coefficients}. Several areas show significant positive effects relative to the baseline (applications): general machine learning ($0.085$, $p = 0.001$), probabilistic methods ($0.132$, $p = 0.002$), and theory ($0.108$, $p = 0.0001$). No areas show significant negative effects.

\begin{table}[h]
\centering
\begin{tabular}{|l|c|c|c|c|}
\hline
\textbf{Area} & \textbf{Coefficient} & \textbf{Std Error} & \textbf{t-statistic} & \textbf{p-value} \\
\hline
Applications (baseline) & 0.000 & --- & --- & --- \\
General Machine Learning & 0.085 & 0.024 & 3.47 & 0.001 \\
Probabilistic Methods & 0.132 & 0.043 & 3.08 & 0.002 \\
Theory & 0.108 & 0.027 & 3.97 & 0.0001 \\
\hline
\end{tabular}
\caption{Significant area coefficients for ICML 2025 regression. Only areas with $p < 0.05$ are shown. All coefficients are relative to the baseline (applications).}
\label{tab:icml2025_area_coefficients}
\end{table}
\subsection{Full Within Paper Paired Analyses}
We present the full results of our within-paper paired analyses below, including every rating field available to us, like presentation, soundness, contribution, quality, clarity, significance and confidence.
\begin{table}[h]
\centering
\small
\begin{adjustbox}{max width=\textwidth}  
\begin{tabular}{|l|c|c|c|c|c|}
\hline
\textbf{Metric} & \textbf{CATE (LLM-aided Paper)} & \textbf{CATE (Human Paper)} & \textbf{Difference} & \textbf{p-value} & \textbf{Effect} \\
\hline
Confidence & $-0.012$ & $-0.116$ & $+0.104$ & \textcolor{green}{0.0073} & Small \\
\hline
Soundness & 0.189 & 0.110 & $+0.080$ & \textcolor{green}{0.0288} & V. Small \\
\hline
Rating & 0.340 & 0.244 & $+0.096$ & \textcolor{orange}{0.0576} & V. Small \\
\hline
Presentation & 0.142 & 0.106 & $+0.036$ & 0.0897 & V. Small \\
\hline
Contribution & 0.155 & 0.117 & $+0.038$ & 0.1888 & V. Small \\
\hline
\end{tabular}
\end{adjustbox}
\caption{Result of Mann-Whitney U test (one-tailed) on ICLR data (all fields) from 2024 and 2025 combined. Kindness difference = LLM rating $-$ Human rating. Total papers: 10,945 (LLM aided papers: 405, Human papers: 10,540).}
\label{tab:iclr_kindness_analysisapp}
\end{table}
\begin{table}[h]
\centering
\small
\begin{adjustbox}{max width=\textwidth}  
\begin{tabular}{|l|c|c|c|c|c|}
\hline
\textbf{Metric} & \textbf{CATE (LLM-aided Paper)} & \textbf{CATE (Human Paper)} & \textbf{Difference} & \textbf{p-value} & \textbf{Effect} \\
\hline
Presentation & 0.215 & 0.038 & $+0.178$ & \textcolor{green}{0.0122} & Small \\
\hline
Soundness & 0.148 & 0.034 & $+0.114$ & 0.0785 & Small \\
\hline
Rating & 0.091 & $-0.029$ & $+0.121$ & 0.1617 & Small \\
\hline
Contribution & 0.111 & 0.044 & $+0.067$ & 0.3070 & V. Small \\
\hline
Confidence & $-0.105$ & $-0.101$ & $-0.004$ & 0.3963 & V. Small \\
\hline
\end{tabular}
\end{adjustbox}
\caption{Result of Mann-Whitney U test (one-tailed) on ICLR data (accepted papers only) from 2024 and 2025 combined. Kindness difference = LLM rating $-$ Human rating. Total papers: 3,543 (LLM aided papers: 94, Human papers: 3,449).}
\label{tab:iclr_kindness_analysis_acceptedapp}
\end{table}
\begin{table}[h]
\centering
\small
\begin{adjustbox}{max width=\textwidth}  
\begin{tabular}{|l|c|c|c|c|c|}
\hline
\textbf{Metric} & \textbf{CATE (LLM-aided Paper)} & \textbf{CATE (Human Paper)} & \textbf{Difference} & \textbf{p-value} & \textbf{Effect} \\
\hline
Contribution & 0.336 & 0.062 & $+0.274$ & \textcolor{green}{0.0258} & Small \\
\hline
Soundness & 0.216 & 0.059 & $+0.157$ & 0.0798 & Small \\
\hline
Rating & 0.310 & 0.041 & $+0.269$ & 0.1428 & Small \\
\hline
Presentation & 0.129 & 0.057 & $+0.072$ & 0.2668 & V. Small \\
\hline
Confidence & $-0.261$ & $-0.081$ & $-0.180$ & 0.8612 & Small \\
\hline
\end{tabular}
\end{adjustbox}
\caption{Result of Mann-Whitney U test (one-tailed) on NeurIPS 2024 data. Kindness difference = LLM rating $-$ Human rating.  Total papers: 2,285 (LLM aided papers: 29, Human papers: 2,256).}
\label{tab:neurips2024_kindness_analysisapp}
\end{table}
\begin{table}[h]
\centering
\small
\begin{adjustbox}{max width=\textwidth}  
\begin{tabular}{|l|c|c|c|c|c|}
\hline
\textbf{Metric} & \textbf{CATE (LLM-aided Paper)} & \textbf{CATE (Human Paper)} & \textbf{Difference} & \textbf{p-value} & \textbf{Effect} \\
\hline
Significance & 0.254 & 0.072 & $+0.182$ & \textcolor{green}{0.0315} & Small \\
\hline
Rating & 0.154 & 0.003 & $+0.151$ & 0.0955 & Small \\
\hline
Quality & 0.139 & 0.071 & $+0.068$ & 0.0997 & V. Small \\
\hline
Clarity & 0.110 & 0.075 & $+0.035$ & 0.2380 & V. Small \\
\hline
Originality & 0.164 & 0.097 & $+0.067$ & 0.2725 & V. Small \\
\hline
Confidence & $-0.005$ & $-0.058$ & $+0.053$ & 0.3476 & V. Small \\
\hline
\end{tabular}
\end{adjustbox}
\caption{Result of Mann-Whitney U test (one-tailed) on NeurIPS 2025 data. Kindness difference = LLM rating $-$ Human rating. Total papers: 2,697 (LLM aided papers: 66, Human papers: 2,631).}
\label{tab:neurips2025_kindness_analysisapp}
\end{table}
\subsection{Full Synthetic Within Paper Paired Analyses}
\label{sec:appwithin}
\begin{table}[H]
\centering
\small
\begin{adjustbox}{max width=\textwidth}  
\begin{tabular}{|l|c|c|c|c|c|}
\hline
\textbf{Metric} & \textbf{CATE (LLM-aided Paper)} & \textbf{CATE (Human Paper)} & \textbf{Difference} & \textbf{p-value} & \textbf{Effect} \\
\hline
Rating & 2.248 & 1.623 & $+0.625$ & \textcolor{green}{0.0018} & Large \\
\hline
Presentation & 0.771 & 0.479 & $+0.292$ & \textcolor{green}{0.0001} & Small \\
\hline
Contribution & 1.521 & 1.274 & $+0.246$ & \textcolor{green}{0.0024} & Small \\
\hline
Soundness & 1.057 & 0.960 & $+0.097$ & 0.1383 & V. Small \\
\hline
Confidence & 0.142 & 0.359 & $-0.217$ & 0.9958 & Small \\
\hline
\end{tabular}
\end{adjustbox}
\caption{Result of Mann-Whitney U test (one-tailed) on ICLR 2025 data with synthetic (fully LLM-generated) reviews. Kindness difference = LLM-generated score $-$ average human score. Total papers: 540 (LLM aided papers: 60, Human papers: 480).}
\label{tab:iclr_synthetic_kindness_analysisapp}
\end{table}
\begin{table}[H]
\centering
\small
\begin{adjustbox}{max width=\textwidth}  
\begin{tabular}{|l|c|c|c|c|c|}
\hline
\textbf{Metric} & \textbf{CATE (LLM-aided Paper)} & \textbf{CATE (Human Paper)} & \textbf{Difference} & \textbf{p-value} & \textbf{Effect} \\
\hline
Rating & 0.674 & 0.712 & $-0.038$ & 0.5670 & V. Small \\
\hline
Confidence & 0.252 & 0.450 & $-0.198$ & 0.9978 & Small \\
\hline
Soundness & 0.689 & 0.772 & $-0.083$ & 0.9205 & V. Small \\
\hline
Presentation & 0.321 & 0.322 & $-0.001$ & 0.7216 & V. Small \\
\hline
Contribution & 1.037 & 1.052 & $-0.015$ & 0.5890 & V. Small \\
\hline
\end{tabular}
\end{adjustbox}
\caption{Result of Mann-Whitney U test (one-tailed) on ICLR 2025 data with synthetic (fully LLM-generated) reviews, accepted papers only. Kindness difference = LLM-generated score $-$ average human score. Total papers: 375 (LLM aided papers: 75, Human papers: 300).}
\label{tab:iclr_synthetic_accepted_kindness_analysisapp}
\end{table}
\begin{table}[H]
\centering
\small
\begin{adjustbox}{max width=\textwidth}  
\begin{tabular}{|l|c|c|c|c|c|}
\hline
\textbf{Metric} & \textbf{CATE (LLM-aided Paper)} & \textbf{CATE (Human Paper)} & \textbf{Difference} & \textbf{p-value} & \textbf{Effect} \\
\hline
Quality & 0.636 & 0.519 & $+0.117$ & \textcolor{green}{0.0468} & Small \\
\hline
Clarity & 0.391 & 0.272 & $+0.119$ & \textcolor{green}{0.0474} & Small \\
\hline
Significance & 0.841 & 0.776 & $+0.065$ & 0.2042 & V. Small \\
\hline
Originality & 0.879 & 0.831 & $+0.048$ & 0.2634 & V. Small \\
\hline
Rating (Overall) & 0.408 & 0.383 & $+0.025$ & 0.3616 & V. Small \\
\hline
Confidence & 0.311 & 0.430 & $-0.119$ & 0.9619 & Small \\
\hline
\end{tabular}
\end{adjustbox}
\caption{Result of Mann-Whitney U test (one-tailed) on NeurIPS 2025 data with synthetic (fully LLM-generated) reviews. Kindness difference = LLM-generated score $-$ average human score. Total papers: 375 (LLM aided papers: 75, Human papers: 300).}
\label{tab:neurips_synthetic_kindness_analysisapp}
\end{table}

\subsection{Generating Fully LLM Reviews}
\label{sec:appprompts}
The prompts used to generate reviews for each of the three conferences are provided below (ICLR: Figure \ref{fig:iclr_synthetic_review_prompt}, ICML: Figure \ref{fig:icml_synthetic_review_prompt}, NeurIPS: Figure \ref{fig:neurips_synthetic_review_prompt}). The reviewer guidelines for each conference are taken directly from the official conference websites (see \citet{iclr2025reviewerguide, neurips2025reviewerguidelines, icml2025reviewerinstructions}).
\begin{figure}[h]
\centering
\fbox{%
\begin{minipage}{0.9\linewidth}
\small
\ttfamily
You are an expert reviewer for a top machine learning conference.

A PDF research paper has been attached above. Read the PDF file completely and provide your assessment following the guidelines below. This is a one-shot task and the user will not be able to provide you with any more information.

\{iclr\_reviewer\_guidelines\}

Review Template: 

Provide the review in JSON format with the following fields in order:

1. Summary of the Paper

2. Strengths of the Paper

3. Weaknesses of the Paper

4. Questions for Authors

5. Flag for ethics review: (Yes or no)

6. Presentation: (a score from 1 to 4)

7. Contribution: (a score from 1 to 4)

8. Soundness: (a score from 1 to 4)

9. Rating: A final rating for the paper in accordance with the following rating scale: \{iclr\_rating\_scale\}

10. Confidence: Your confidence in the rating you provided in accordance with the following confidence scale: \{iclr\_confidence\_scale\}

This JSON will be automatically parsed, so ensure the format is precise. Do not include numerals in the JSON keys.
\end{minipage}%
}
\caption{Prompt used to generate ICLR reviews}
\label{fig:iclr_synthetic_review_prompt}
\end{figure}
\begin{figure}[h]
\centering
\fbox{%
\begin{minipage}{0.9\linewidth}
\small
\ttfamily
You are an expert reviewer for a top machine learning conference.

A PDF research paper has been attached above. Please read the PDF file completely and provide your assessment following the guidelines below. This is a one-shot task and the user will not be able to provide you with any more information.

\{icml\_reviewer\_guidelines\}

Review Template:

Provide the review in JSON format with the following fields in order, as described in the reviewer guidelines above:

1. Summary

2. Claims and Evidence

3. Methods and Evaluation Criteria

4. Theoretical Claims

5. Experimental Designs or Analyses

6. Supplementary Material

7. Relation To Broader Scientific Literature

8. Essential References Not Discussed

9. Other Strengths and Weaknesses

10. Other Comments or Suggestions

11. Questions for Authors

12. Code of conduct acknowledgement

13. Overall Recommendation (use JSON field name "Rating")

This JSON will be automatically parsed, so ensure the format is precise. Do not include numerals in the JSON keys.
\end{minipage}%
}
\caption{Prompt used to generate ICML reviews}
\label{fig:icml_synthetic_review_prompt}
\end{figure}
\begin{figure}[h]
\centering
\fbox{%
\begin{minipage}{0.9\linewidth}
\small
\ttfamily
You are an expert reviewer for a top machine learning conference.

A PDF research paper has been attached above. Read the PDF file completely and provide your assessment following the guidelines below. This is a one-shot task and the user will not be able to provide you with any more information.

\{neurips\_reviewer\_guidelines\}

Review Template:

Provide the review in JSON format with the following fields in order, as described in the reviewer guidelines above:

1. Summary

2. Strengths and Weaknesses

3. Quality

4. Clarity

5. Significance

6. Originality

7. Questions

8. Limitations

9. Overall

10. Confidence

11. Ethical concerns

12. Code of conduct acknowledgement

This JSON will be automatically parsed, so ensure the format is precise. Do not include numerals in the JSON keys.
\end{minipage}%
}
\caption{Prompt used to generate NeurIPS reviews}
\label{fig:neurips_synthetic_review_prompt}
\end{figure}

\begin{figure}[h]
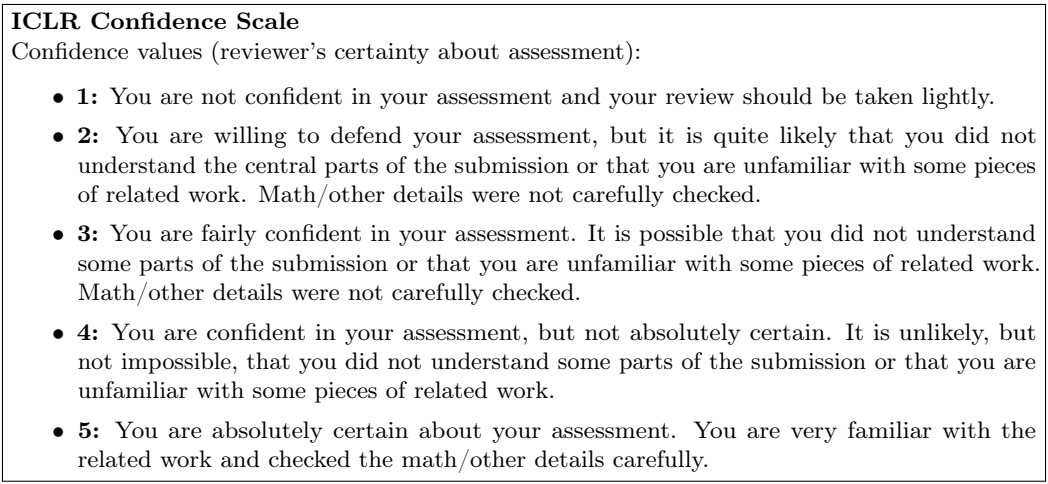

\centering
\fbox{%
\begin{minipage}{0.9\linewidth}
\small
\textbf{ICLR Confidence Scale}

Confidence values (reviewer's certainty about assessment):

\begin{itemize}
    \item \textbf{1:} You are not confident in your assessment and your review should be taken lightly.
    
    \item \textbf{2:} You are willing to defend your assessment, but it is quite likely that you did not understand the central parts of the submission or that you are unfamiliar with some pieces of related work. Math/other details were not carefully checked.
    
    \item \textbf{3:} You are fairly confident in your assessment. It is possible that you did not understand some parts of the submission or that you are unfamiliar with some pieces of related work. Math/other details were not carefully checked.
    
    \item \textbf{4:} You are confident in your assessment, but not absolutely certain. It is unlikely, but not impossible, that you did not understand some parts of the submission or that you are unfamiliar with some pieces of related work.
    
    \item \textbf{5:} You are absolutely certain about your assessment. You are very familiar with the related work and checked the math/other details carefully.
\end{itemize}
\end{minipage}%
}
\caption{ICLR confidence scale used in synthetic review generation}
\label{fig:confidencescale}
\end{figure}

The prompt used to generate synthetic metareviews is provided in Figure~\ref{fig:iclr_synthetic_metareview_prompt}. Here, iclr$\_$rating$\_$scale and iclr$\_$confidence$\_$scale refer to the set of possible ratings and confidence scores reviewers can assign and their given interpretations (for example: 5: Weak Accept). These are provided in detail in Figure \ref{fig:ratingscale} and Figure \ref{fig:confidencescale}.
\begin{figure}[h]
\centering
\fbox{%
\begin{minipage}{0.9\linewidth}
\small
\ttfamily
You are an area chair for a top machine learning conference.

You have been provided with all the reviews for a paper submission. Your task is to make a final accept/reject decision based on these reviews.

\#\# Rating Scale for Reference:
\{iclr\_rating\_scale\}

\#\# Confidence Scale for Reference:
\{iclr\_confidence\_scale\}

Presentation, Contribution, and Soundness are all scores from 1 to 4.

\#\# Reviews:
\{\{reviews\}\}

\#\# Your Task:
Based on the reviews above, make a final decision for this paper.

Provide your response in JSON format with exactly one field:
1. "decision": Either 1 (Accept) or 0 (Reject) - use integer values only

Ensure the JSON is valid and parseable. Do not include any text outside the JSON object.
\end{minipage}%
}
\caption{Prompt used to generate ICLR decisions (metareview)}
\label{fig:iclr_synthetic_metareview_prompt}
\end{figure}
\begin{figure}[h]
\centering
\fbox{%
\begin{minipage}{0.9\linewidth}
\small
\textbf{ICLR Rating Scale}

\begin{itemize}
    \item \textbf{1:} Strong Reject
    \item \textbf{3:} Reject, not good enough
    \item \textbf{5:} Marginally below the acceptance threshold
    \item \textbf{6:} Marginally above the acceptance threshold
    \item \textbf{8:} Accept, good paper
    \item \textbf{10:} Strong accept, should be highlighted at the conference
\end{itemize}
\end{minipage}%
}
\caption{ICLR rating scale used in synthetic review generation}
\label{fig:ratingscale}
\end{figure}

\end{document}